\documentclass[journal]{IEEEtran}
\usepackage{blindtext,graphicx,booktabs,amsmath,amssymb,tabularx,tabu,mathrsfs}
\usepackage[ruled]{algorithm2e}
\usepackage{color, colortbl}
\usepackage{balance}
\usepackage{xcolor}
\usepackage{makecell}
\usepackage{subfigure}
\usepackage{multirow}

\newcommand{\frenet}{Frenét }

\makeatletter
\def\endthebibliography{%
  \def\@noitemerr{\@latex@warning{Empty `thebibliography' environment}}%
  \endlist
}
\makeatother

\begin{document}
\title{Dynamically Conservative Self-Driving Planner for Long-Tail Cases}
\author{Weitao Zhou, Zhong Cao, Nanshan Deng, Xiaoyu Liu, Kun Jiang, Diange Yang

\thanks{All authors are with the School of Vehicle and Mobility, Tsinghua University, Beijing, China 100084.
 {\tt\small \{zwt19,dns18,xiaoyu-l21\}@mails.tsinghua.edu.cn;}  {\tt\small\{caozhong,jiangkun,ydg\}@tsinghua.edu.cn;}

Z. Cao, D. Yang are the corresponding authors.
}  
}

\markboth{IEEE TRANSACTIONS ON INTELLIGENT TRANSPORTATION SYSTEMS}%
{Shell \MakeLowercase{\textit{et al.}}: Bare Demo of IEEEtran.cls for Journals}

\maketitle

\begin{abstract}
Self-driving vehicles (SDVs) are becoming reality but still suffer from  ``long-tail" challenges during natural driving: the SDVs will continually encounter rare, safety-critical cases that may not be included in the dataset they were trained. Some safety-assurance planners solve this problem by being conservative in all possible cases, which may significantly affect driving mobility. To this end, this work proposes a method to automatically adjust the conservative level according to each case's ``long-tail” rate, named dynamically conservative planner (DCP). We first define the ``long-tail" rate as an SDV's confidence to pass a driving case. The rate indicates the probability of safe-critical events and is estimated using the statistics bootstrapped method with historical data. Then, a reinforcement learning-based planner is designed to contain candidate policies with different conservative levels. The final policy is optimized based on the estimated ``long-tail” rate. In this way, the DCP is designed to automatically adjust to be more conservative in low-confidence ``long-tail” cases while keeping efficient otherwise.
The DCP is evaluated in the CARLA simulator using driving cases with ``long-tail” distributed training data. The results show that the DCP can accurately estimate the ``long-tail" rate to identify potential risks. Based on the rate, the DCP automatically avoids potential collisions in ``long-tail" cases using conservative decisions while not affecting the average velocity in other typical cases. Thus, the DCP is safer and more efficient than the baselines with fixed conservative levels, e.g., an always conservative planner. 
This work provides a technique to guarantee SDV’s performance in unexpected driving cases without resorting to a global conservative setting, which contributes to solving the ``long-tail" problem practically.

\end{abstract}

 \begin{IEEEkeywords}
Self-driving Vehicle, Reinforcement Learning, Trajectory Planning, Long-Tail
 \end{IEEEkeywords}

\IEEEpeerreviewmaketitle

\section{Introduction}

Self-driving vehicles (SDVs) have undergone significant progress recently. The state-of-art SDVs can achieve uninterrupted autonomous driving for tens of thousands of mileages without human disengagement \cite{schwall2020waymo}. 
However, the current SDVs are still impractical due to safety concerns on ``long-tail'' cases, which commonly refer to the rare and endless safe-critical driving cases an SDV may encounter in real-world driving \cite{jain2021autonomy}. 
These cases may have various characteristics, e.g., road closures, road accidents, and agents violating traffic rules, which cannot be enumerated even after millions of testing miles on public roads \cite{kalra2016driving}. 
As such, the SDVs may still encounter unfamiliar but risky driving cases after being elaborately designed.
This phenomenon raises significant concerns and should be addressed before SDVs are widely applied. 
Notably, the ``long-tail'' problem may originate from different SDV components \cite{anguelov2019taming}; we focus on the planning module in this study. Our goal is to design a planner to make performance-guaranteed decisions under ``long-tail'' cases, which may be safe-critical but rarely or even not contained in the dataset that trains or tests the planner.

Related methods to consider ``long-tail'' problems for planners include: 1) limiting the operation domain of the planner, 2) improving the planner’s performance to cover more scenarios, and 3) modifying the planner to be globally conservative.

A straightforward method to avoid ``long-tail'' problems is not using the self-driving function in driving cases that were not considered while designing the SDVs.
The operational design domain (ODD) concept specifies the scenarios where SDVs are designed to operate safely \cite{society2018sae}. 
The ODD scenarios can be described by specifying driving environment conditions, such as traffic flow density \cite{koopman2019many} and SDV's velocity  \cite{sun2021acclimatizing}. Some methods learn ODD conditions from collected real-world accident/disengagement data \cite{lee2020identifying}.
Besides specifying ODD conditions, other risk-assessment methods are related to this topic \cite{guo2019safe}. Although not yet specifically designed for “long-tail” risks, existing works define driving risks from different aspects, such as driving scenario status, events features and potential collisions \cite{chia2022risk}. 
A representative metric is Time-To-Collision (TTC), which quantifies the time remaining before a collision occurs \cite{wachenfeld2016worst}. 
Moreover, some data-driven methods can extract potential risky driving features with failure datasets \cite{strickland2018deep}.
The potential safe-critical scenarios for an SDV system can be generated through adversarial learning \cite{ding2022survey}. 
After risks are identified, the SDV can switch to a backup policy, e.g., using decision tree methods \cite{li2017explicit} or considering risks during the planning process \cite{zhou2022long}. These methods can predefine the features of ``long-tail'' scenarios or learn from collected safe-critical events. However, the idea is limited to known or already seen safe-critical cases. Consequently, SDVs may still fail in unexpected or first-time encountered ``long-tail''  cases. 

Some methods improve planner performance to cover more potential ``long-tail'' cases. A well-known method is to train a driving policy using collected ``long-tail'' data, e.g., based on reinforcement learning (RL) or imitation learning \cite{zhu2021survey}. However, a learning-based agent commonly requires enormous training data for each driving case, which may be impractical for a low-probability case. Consequently, the trained model still performs poorly in rare ``long-tail'' cases  \cite{zhang2021deep}.  
Though not explored in planning tasks, a well-known deep learning technique is to re-sample ``long-tail'' case data to emphasize these cases more during training \cite{he2009learning} \cite{kang2019decoupling}. Other methods perform data augmentation in  ``long-tail'' cases using data mixup \cite{zhong2021improving} or Gaussian prior \cite{zang2021fasa}. 
Recently, transfer learning \cite{kiran2021deep} and few-shot learning  \cite{wang2020generalizing} have been proposed to extract knowledge from common driving cases to improve performance in ``long-tail'' cases.
These methods try to reduce the amount of training data and improve the trained model's performance in ``long-tail'' cases. However, they still cannot guarantee performance in unexpected ``long-tail'' cases that were not contained in the training dataset.

Recent research on safe-assurance planners theoretically guarantees performance in all possible cases by designing stricter global constraints. For example, a planner can be modified to be more conservative by enlarging the space occupied by the SDV's surrounding obstacles, which reduces risks from potential unexpected behaviors \cite{xu2014motion} \cite{li2021prediction}. 
Reachability-set-based planners aim to avoid all possible future of surrounding objects to guarantee safety  \cite{ahn2020reachability} \cite{althoff2021set}. 
The set-based verification techniques can also be used to design planners that guarantee the ``legal safety" of SDVs \cite{pek2020using}. Moreover, the responsibility sensitive safety (RSS) method uses a formal model to determine if the output of a planner causes responsible collisions or accidents \cite{shalev2017formal}.
Such planners satisfy safety constraints through formal verification to guarantee performance in any driving scenario, including the untrained/unexpected ``long-tail'' cases. 
However, the strict constraints (e.g., considering reachable set) may yield unnecessarily over-conservative decisions in some typical cases, such as car-following, which significantly affects driving mobility.

In this study, we aim to design a planner that guarantees performance in unexpected ``long-tail'' cases but does not resort to a global conservative setting. Instead, the planner will identify ``long-tail'' cases and adjust to be conservative accordingly.
To achieve the goal, we first formalize the ``long-tail'' problem in SDV planning as a driving confidence problem. Then, we define the ``long-tail'' rate to quantify how confident an SDV's policy is to pass a case. 
The confidence is calculated statistically and designed to be low in potential safe-critical cases, including the unexpected cases that are not contained in the training dataset. 
Then an RL-based planner is designed to automatically adjust the conservative level according to the ``long-tail'' rate in different cases. The planner will make conservative decisions to guarantee performance in potential safe-critical cases and maintain efficiency in high-confidence cases. The overall framework is called dynamically conservative planner (DCP).

The detailed contributions of this study are as follows.

1)  Dynamically Conservative Planner (DCP) Framework: The DCP framework is based on model-based RL containing candidate policies with different conservative levels. The DCP will switch to different policies according to each encountered case's estimated ``long-tail” rate. 

2) ``Long-tail'' Rate Estimation: We defined the ``long-tail'' rate as the driving confidence for an encountered driving case. This confidence is designed to be low in ``long-tail''  cases and high in other cases. It is estimated based on statistics bootstrapped methods using historical driving data. 

3) Dynamically Conservative Policy Generation: We design the DCP to choose a more conservative policy when the encountered case has a higher ``long-tail'' rate (lower confidence). As such, the DCP automatically adjusts to be conservative in low-confidence cases and maintains efficiency in high-confidence cases.

The remainder of this article is organized as follows. Section II introduces the preliminaries and formally defines the ``long-tail'' problem. The proposed DCP method is presented in Section III. Section IV presents a case study and results. Finally, Section V concludes this study.

\section{Preliminaries and Problem Setup}

\subsection{Preliminaries}

Following \cite{kochenderfer2015decision}, the planning problem of SDVs can be formulated as a Markov decision process (MDP) or partially observation Markov decision process (POMDP).
The MDP assumption makes the problem satisfy the Markov property: the conditional probability distribution of future states of the process depends only on the present state. 
In general, the agent (SDV) should optimize long-term rewards under a sequential decision-making setting. 

In detail, an MDP can be defined by the tuple $(\mathcal{S}, \mathcal{A}, \mathcal{R}, \mathcal{T})$ comprising:

\begin{itemize}
\item a state space $\mathcal{S}$
\item an action space $\mathcal{A}$
\item a reward function $\mathcal{R}$
\item a transition operator (probability distribution) $\mathcal{T}$: $\mathcal{S}\times\mathcal{A}\times \mathcal{S}\to \mathbb{R}$
\item a discount factor $\gamma\in(0,1]$ is set as a fixed value to favor immediate rewards over rewards in the future.
\item the planning horizon is defined as $H\in\mathbb{N}$. 
\end{itemize}

A general policy $\pi\in\Pi$ maps each state to a distribution over actions.  $\Pi$ denotes the set containing all candidate policies $\pi$. Namely, $\pi(a|s)$ denotes the probability of taking action $a$ in state $s$ using the policy $\pi$. 

During the driving process, the SDV should make a decision action $a$ (following the policy $\pi$) with the observed state $s$  at each timestep; then, it observes the next state $s'$ and reward $r$ from the environment. Starting from state $s$ and following a policy $\pi$, the collected data for $H$ future time steps can be represented in the form of trajectory $\omega_{\pi}(s)$:

\begin{equation}
\omega_{\pi}(s):=\left\{s_0, a_0, r_0, s_1, a_1, r_1\ldots, s_{H}, a_{H}, r_{H}\right\}
\label{trajectory}
\end{equation}

The driving performance of a policy $\pi$ can be described as value $V_{\pi}$ and action-value functions $Q_{\pi}$. For a given reward function $r$, these functions are defined as follows:
\begin{equation}
      V_{\pi}(s)=\mathbb{E}_{\pi}[\sum_{t=0}^{H} \gamma^{t} r_t|s_0=s] \\
\label{equ:Vdefine}
\end{equation}
\begin{equation}
      Q_{\pi}(s,a_{\pi})=\mathbb{E}_{\pi}[\sum_{t=0}^{H}  \gamma^{t} r_t|s_0=s,a_0=a_{\pi}]
\label{equ:Qdefine}
\end{equation}
where $t=0,1,2,3...$ denotes the discrete time steps. For each time step $t$, $a_t\thicksim \pi(a_t|s_t),s_{t+1} \thicksim T(s_{t+1}|s_t,a_t)$. $\mathbb{E}[\cdot]$ denotes the expectation with respect to this distribution. 

Given the policy space $\Pi$, the goal of an SDV's planner is to find the optimal policy $\pi^{*}\in\Pi$, which has the maximum driving performance (value) for all $s \in \mathcal{S}$:

\begin{equation}
\pi^{*}(s)=\underset{\pi\in\Pi}{\operatorname{argmax}} V_{\pi}(s)
\end{equation}

\begin{figure}[htbp]
    \subfigure[Typical Case]{
    \begin{minipage}[t]{0.49\linewidth}
    \centering
    \includegraphics[width=\linewidth]{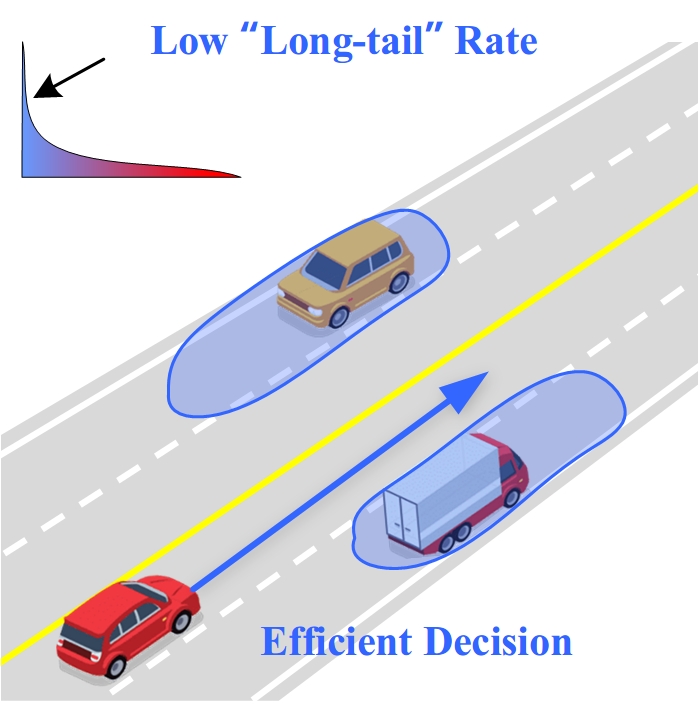}
    \end{minipage}%
    }%
    \subfigure[``Long-Tail'' Case]{
      \begin{minipage}[t]{0.49\linewidth}
      \centering
      \includegraphics[width=\linewidth]{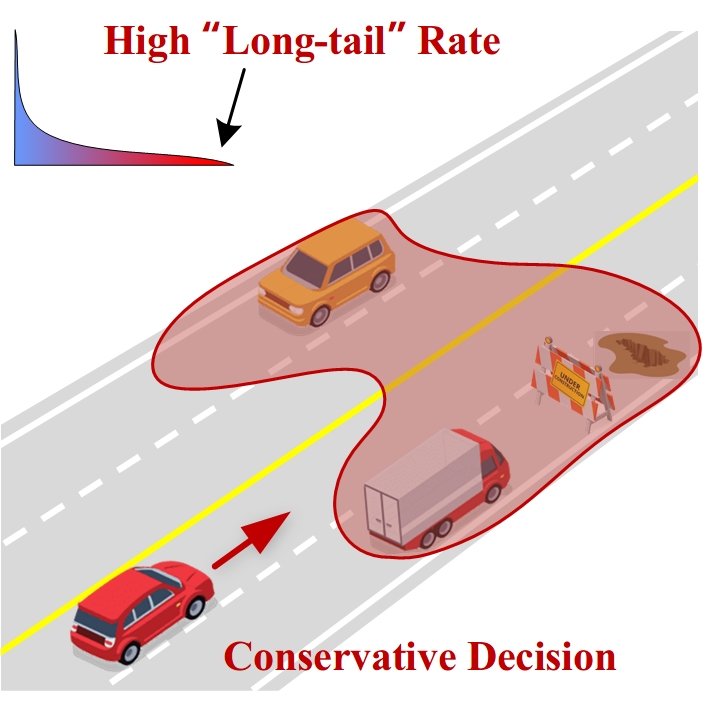}
      \end{minipage}%
      }%
    \caption{The dynamically conservative planning problem. The planner will: 1) estimate the ``long-tail'' rate of an encountered driving case and 2) automatically adjust the conservative level of decision. The DCP will make efficient decisions in typical cases and conservative decisions in  ``long-tail'' cases.}
\label{fig:problem}
\end{figure}
\subsection{Problem Setup}

This study focuses on designing a planner to automatically adapt to unexpected ``long-tail'' cases. The proposed DCP method aims to be conservative in ``long-tail'' cases while maintaining efficiency in typical cases (Fig.  \ref{fig:problem}). Particularly, there are two problems. First, the ego SDV will estimate the  \emph{``long-tail'' rate} of an encountered driving case, which describes the policy's confidence of passing the case. 
This rate is determined by the confidence of the surrounding agents' transition estimation. 
When the training data is sufficient to estimate the transition accurately in typical cases, the confidence of transition can be represented as a narrow undrivable area that captures agent behavior accurately (blue area, left).
On the other hand, a low-confidence estimation in the rare ``long-tail'' case leads to a larger area to cover unexpected possibilities (red area, right). 
Then, the planner adjusts the conservative level accordingly in different cases, defined as \emph{dynamically conservative planning process}. The overall DCP process can be described as follows:

\begin{equation}
a_{dc} = R_d(s, L(s, \Pi, D(s)))
\label{planning_process}
\end{equation}
where $R_d$ denotes the DCP process, $a_{dc}$ represents the dynamically conservative action, $D(s)$ denote the collected historical training data on encountered driving case $s$, $\Pi$ denotes candidate policy group, and $L(s, \Pi, D(s))$ denotes the driving policies' ``long-tail'' rate estimation process. In this study, we assume that a conservative planner considering a reachable set can avoid an SDV's failures in unexpected ``long-tail” cases \cite{althoff2021set}. The cases with inevitable accidents are excluded, i.e., all possible actions of SDV cannot avoid a collision.

\section{Method}

\begin{figure}[ht]
	\centering
	\includegraphics[width=\linewidth]{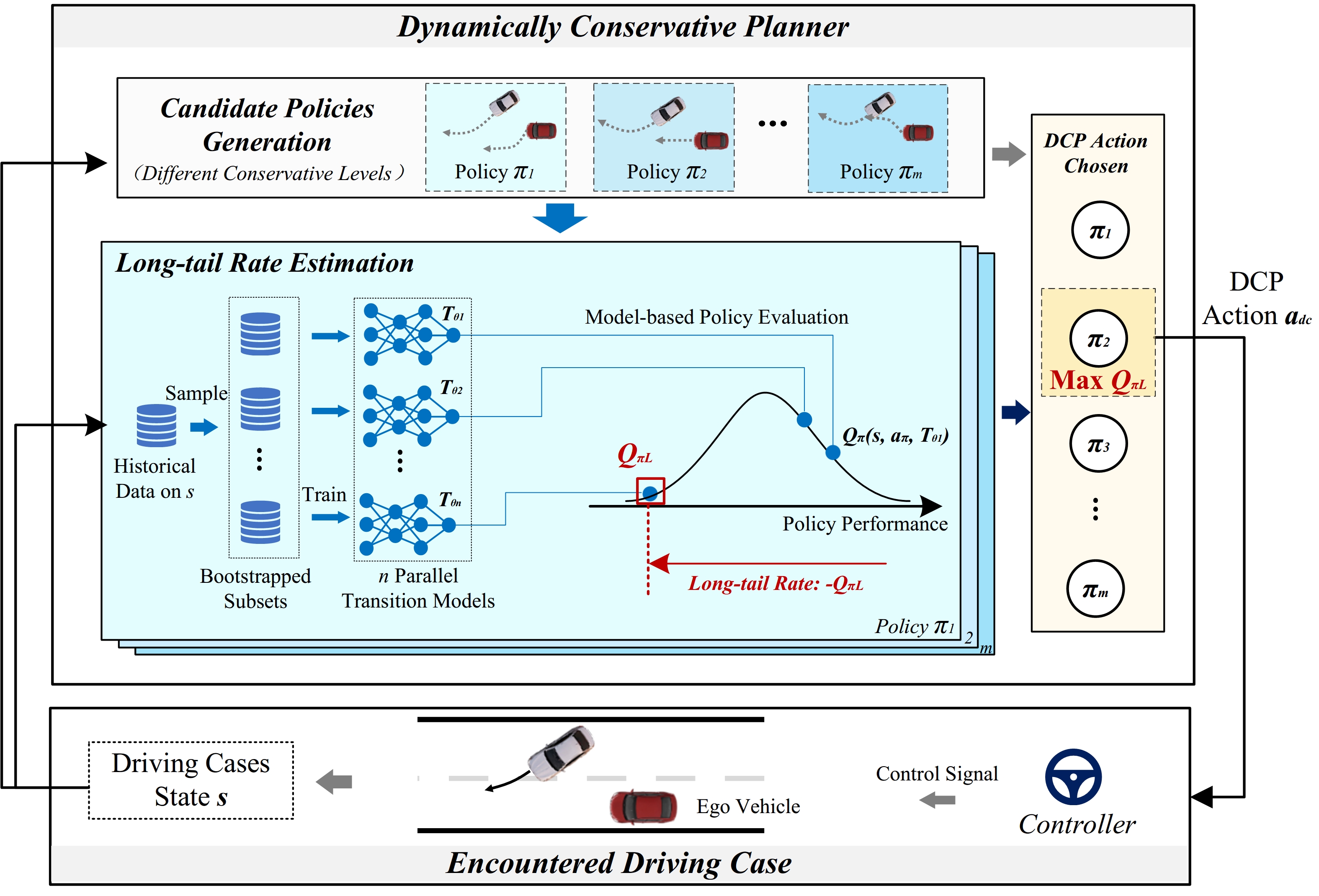}
	\caption{
    The proposed DCP framework. It first generates candidate policies with different conservative levels for an encounter driving case $s$. Then each policy’s ``long-tail” rate is estimated using ensemble models. The ensemble structure contains $n$ parallel transition models, trained with sampled bootstrapped subsets respectively, to indicate the confidence of transition training. Each transition model can be used to evaluate policy performance. The “long-tail” rate is the minimum estimated performance $Q_{\pi L}$. A potential low “long-tail” rate value indicates potential safe-critical events. Finally, the DCP will choose the action with the highest “long-tail” rate value to automatically adjust the conservative level. }
	\label{fig:framework}
\end{figure}

The proposed DCP framework is shown in Fig.  \ref{fig:framework}. The DCP receives state $s$ from an encountered driving case and output decision as DCP action $a_{dc}$. During the planning process, the DCP first generates several candidate policies with different conservative levels. Then the DCP estimates the ``long-tail'' rate of each policy. The DCP generates the final action with the maximum ``long-tail'' rate value. 

In this section, we first introduce how the ``long-tail'' rate for a policy is defined and derived theoretically from driving confidence. Then, we describe the estimation of confidence using bootstrapped methods based on the model-based RL method. Afterward, the method of candidate policy generation and dynamically conservative action generation is introduced.

\subsection{``Long-Tail'' Rate Estimation}

\subsubsection{``Long-Tail'' Rate Definition}

The ``long-tail'' rate is designed to identify unexpected cases that might lead to failures, which is defined based on driving confidence that satisfies the following:
\begin{equation}
\begin{split}
&\forall \pi \in \Pi, 
 P\left(\tilde{Q}_{\pi}(s, a_{\pi}) \geq Q_{\pi L}(s, a_{\pi}, D) |D\right) \geq 1-\delta
\end{split}
\label{long_tail_rate}
\end{equation}
where $Q_{\pi L}(s, a_{\pi}, D)$ defines the confidence of a policy $\pi$ applied to driving case $s$, $D$ denotes the historical collected dataset, $\tilde{Q}_{\pi}(s, a_{\pi}) $ denotes the true value under the policy $\pi(s)$, $\delta \in (0,1)$ is a small value to constraint the probability that  $\tilde{Q}_{\pi}(s, a_{\pi}) $ is greater than the $Q_{\pi L}(s, a_{\pi}, D)$. 

Eq. \eqref{long_tail_rate} defines that  $Q_{\pi L}(s, a_{\pi}, D)$ should be a probabilistic lower bound of true driving policy performance $\tilde{Q}_{\pi}(s, a_{\pi}) $ with $D$. A higher $Q_{\pi L}(s, a_{\pi}, D)$ indicates the policy is confident that it will achieve good performance, hence is less likely to lead to unexpected failures. Conversely, a low  $Q_{\pi L}(s, a_{\pi}, D)$ means that we cannot gain enough confidence to pass the case according to historical data, indicating potential safe-critical events.

Notably, a higher ``long-tail” rate $L_{\pi} (s, D)$ indicates a higher probability of safe-critical events, referring to a lower $Q_{\pi L}(s, a_{\pi}, D)$ value:
\begin{equation}
L_{\pi} (s,D)=-Q_{\pi L}(s, a_{\pi}, D)
\end{equation}

\subsubsection{Driving Policy Performance Estimation}

The driving policy performance ${Q_{\pi}}\left(s, a_{\pi}\right)$ (defined in Eq. \eqref{equ:Qdefine}) is evaluated using a model-based RL method called model value expansion \cite{feinberg2018model}. The main idea is to ``imagine'' the policy's outcomes using a trained transition model of the environment.

Particularly, the transition model enables the generation of imaginary trajectory started from $s$ following policy $\pi$: 
\begin{equation}
\hat{\omega_{\pi}}(s):=\left\{s_0, \pi(s_0), \hat{r_0}, \hat{s}_1, \pi(\hat{s}_1), \hat{r}_1\ldots, \hat{s}_t, \pi(\hat{s}_t), \hat{r}_t\right\}
\label{imaginary_trajectory}
\end{equation}
where $\hat{s}_{t+1} =\hat{T}(s_t, \pi(s_t))$ denotes the one-step imagination, determined by the transition model $\hat{T}$ and policy $\pi$.  $\hat{r}_{t}$ denotes rewards determined by a predefined reward function $R$:
\begin{equation}
\hat{r}_t = R(s_t, a_t)
\end{equation}

For one imaginary trajectory $\hat{\omega_{\pi}}(s)$, its cumulative driving reward (in horizon $H$) can be written as follows:
\begin{equation}
G(\hat{\omega_{\pi}}(s))=\sum_{t=0}^{H}\gamma^{t} \hat{r}_{t},  \hat{r}_{t} \in  \hat{\omega_{\pi}}(s), s_{0}=s
\end{equation}

Given $j$ imaginary trajectories as a dataset $\hat{D}_{\pi}^{j}(s)$, the driving performance of a policy is then estimated by:

\begin{equation}
\hat{D}_{\pi}^j(s)= \{\hat{\omega}_{\pi}^1(s), \hat{\omega}_{\pi}^2(s), ... \hat{\omega}_{\pi}^j(s) \}
\end{equation}

\begin{equation}
\begin{split}
\forall s \in \mathbb{S},  \hat{Q_{\pi}}\left(s, a_{\pi}, \hat{T}\right) \leftarrow &\bar{G}\left(s, a_{\pi}, \hat{T}\right)=\\
&\frac{1}{j} \sum G(\hat{\omega}_{\pi}^j(s)) , \hat{\omega}_{\pi}^j(s)\in \hat{D}_{\pi}^j(s))
\label{mve}
\end{split}
\end{equation}

The reward function is commonly predefined in SDV's planning tasks \cite{zhu2021survey}. Thus, the trajectory feedback value $G\left(\hat{\omega_{\pi}}(s)\right)$ at a specific state $s$ is an independent identically distributed (i.i.d) sampling according to the Markov property:
\begin{equation}
\begin{split}
G\left(\hat{\omega_{\pi}}(s)\right) &=\sum_{t=0}^{H} \gamma^{t} \hat{r_{t}} \\
&\sim \sum_{t=0}^{H} \gamma^{t} R\left(\hat{T}\left(s_{t}, \pi\left(s_{t}\right)\right), \pi\left(\hat{T}\left(s_{t}, \pi\left(s_{t}\right)\right)\right)\right) \mid s \\
&=p_{\hat{T}}(s, \pi)
\end{split}
\end{equation}

If the policy $\pi$ is fixed, $G\left(\hat{\omega_{\pi}}(s)\right)$ is i.i.d under the distribution of $p_{\hat{T}} (s,\pi)$.  Thus, the estimated $\bar{G}\left(s, a_{\pi}\right)$ converges to true policy performance $\hat{Q_{\pi}}\left(s, a_{\pi}, \hat{T}\right)$ under transition $\hat{T}$ when the data amount $j$ tend to infinity:
\begin{equation}
{ \lim_{j \to +\infty}\frac{1}{j} \sum G(\hat{\omega_{\pi}}(s)) \rightarrow \hat{Q_{\pi}}\left(s, a_{\pi}, \hat{T}\right)}
\end{equation}

In addition, the estimated driving performance $\bar{G}\left(s, a_{\pi}, \hat{T}\right)$ will approach true policy performance when the estimated transition model $\hat{T}$ approaches the true environmental transition $T$:

\begin{equation}
{ \lim_{j \to +\infty,  \hat{T} \to T}\frac{1}{j} \sum G(\hat{\omega}_{\pi}^k(s)) \rightarrow \tilde{Q_{\pi}}\left(s, a_{\pi}\right)}
\label{mve_converge}
\end{equation}

\subsubsection{Driving Performance Confidence Estimation}

In Eq. \eqref{mve}, the estimated $\bar{G}\left(s, a_{\pi}\right)$ has no guarantee where the transition model $\hat{T}$ is inaccurate. This transition model is commonly trained on collected real-world driving data. We define a collected driving dataset on $s$ as $D_{\pi}^k(s)$, which containing $k$ real-world trajectories:
\begin{equation}
D_{\pi}^k(s)= \{\omega_{\pi}^1(s), \omega_{\pi}^2(s), ... \omega_{\pi}^k(s) \}
\end{equation}

The transition model $\hat{T}$ trained with $D_{\pi}^k(s)$ might be inaccurate in ``long-tail'' driving cases $s$ where the amount of data $k$ is small and even equal to zero, resulting in erroneous driving performance estimation.
 
To quantify how good is the performance estimation, we define a distribution $P\left(\tilde{Q}_{\pi}(s, a_{\pi}) | D_{\pi}^k(s)\right)$. It describes the probability of true value given the current dataset $ D_{\pi}^k(s)$, namely, the confidence of value estimation. Intuitively, the distribution will be concentrated near the true value $\tilde{Q}_{\pi}(s, a_{\pi})$ when the amount of data is large enough for value estimation with high confidence. In addition, the distribution may be more scattered in data-sparse cases where there is not enough confidence in performance estimation \cite{cao2021confidence} \cite{cao2022trustworthy}.

According to the previous section (Eq. \eqref{mve_converge}), the task of estimating driving performance (confidence) can be transformed into the task of estimating transition (confidence):
\begin{equation}
P\left(\tilde{Q}_{\pi}(s, a_{\pi}) | D_{\pi}^k(s)\right) \sim P(\hat{T}(s, a_{\pi})| D_{\pi}^k(s))
\end{equation}

The transition $T$ can be further decomposed in an SDV's planning task. 
Without loss of generality, we consider a state $s$ to contain information about all agents in the environment. More specific, each state $s$ includes the status of ego SDV $s_e$ and $i$ surrounding agents $s_s$.  
\begin{equation}
s_t=\{s^e_{t}, s^s_{t}\}=\{s^e_{t}, s^1_{t},  s^2_{t},... s^i_{t}\}
\end{equation}

We assume that all actions in the action space can only be applied to the SDV (e.g., braking and acceleration). Thus, action $a$ can only affect the state transition of the ego SDV. Therefore, the transition of the entire driving environment can be decomposed as follows:
\begin{equation}
\begin{split}
T\left(s_{t+1} \mid s_{t}, a_{t}\right)=T\left(s^e_{t+1}\mid s^e_{t}, a_{t}\right) T\left(s^s_{t+1} \mid s_{t}\right)
\end{split}
\end{equation}

The first term $T\left(s^e_{t+1}\mid s^e_{t}, a_{t}\right)$ denotes the dynamics of the ego SDV, which can be measured without interaction with the surrounding agents in the environment \cite{kutluay2014validation}. Thus, we assume the DCP already knows the SDV's dynamics model.
The second term $T\left(s^s_{t+1} \mid s_{t}\right)$ denotes the environmental transition, which indicates how the surrounding agents act in a time step. Finally, the driving confidence $P\left(\tilde{Q}_{\pi}(s, a_{\pi}) | D_{\pi}^k(s)\right)$ is determined by the confidence of environmental transition model $T_s\left(s^s_{t+1} \mid s_{t}\right)$:

\begin{equation}
P\left(\tilde{Q}_{\pi}(s, a_{\pi}) | D_{\pi}^k(s)\right) \sim P(\hat{T_s}(s)| D_{\pi}^k(s))
\label{q_t_relationship}
\end{equation}
\subsubsection{``Long-tail'' Rate Estimation}

According to Eq. \eqref{long_tail_rate}, the estimation of the ``long-tail'' rate can be achieved by first estimating the distribution $P\left(\tilde{Q}_{\pi}(s, a_{\pi}) | D\right)$ and then truncating according to the probability constraint $\delta$. Considering Eq. \eqref{long_tail_rate}, if there exists a set of environmental transition models $\mathcal{T}$ that has more than $1-\delta$  probability of containing true environmental transition $\tilde{T}_s$:

\begin{equation}
P(\tilde{T}_s  \in \mathcal{T} | D) \geq 1-\delta
\label{transition_set}
\end{equation}

The ``long-tail'' rate can be calculated as:

\begin{equation}
Q_{\pi L}(s, a_{\pi}, D_{\pi}^k(s))= \arg  \min_{T_s  \in \mathcal{T}}\hat{Q}_{\pi}\left(s, a_{\pi}, T_s\right)
\label{long_tail_rate_set}
\end{equation}

In this study, the set $\mathcal{T}$ is approximated using bootstrapped principle with ensemble models. 
The bootstrapped method infers population properties from sampled data without assumptions about the population distributions. 
A classical bootstrapped principle takes a dataset $D$ and an estimator $\phi$ as inputs. The  $\phi$ will estimate a variable from dataset $D$. To obtain the confidence of such an estimation, a bootstrapped distribution is essential but will not be explicitly calculated. Instead, $n$ bootstrapped subsets $\overline D_1, \overline D_2, ... \overline D_{n}$ will be sampled uniformly with replacement from $D$. Then a sample from the bootstrapped distribution can be calculated using one of the subsets as $\phi(\overline  D)$.

We first build an environmental transition model using a neural network, described as $ T_{\theta}$, which is determined by its parameters $\theta$. Such a neural network is considered powerful enough to fit the true transition matrix $\tilde{T_s}$ when there are sufficient data. The ``true value'' of parameter $\theta$ is defined as $\tilde{\theta}$,  that is, $ T_{\tilde{\theta}}$ approaches $\tilde{T_s}$. We consider the learned parameter $\theta$ as a variable inferred from $D_{\pi}^k(s)$. Then, the confidence of environmental transition estimation can then be seen as the confidence of model parameter estimation $P(\theta, D_{\pi}^k(s))$.

The ensemble network structure contains $n$ transition models $T_{\theta_1}, T_{\theta_2}, ... T_{\theta_n}$ in parallel (Fig.  \ref{fig:framework}). The number of transition models equals the number of bootstrapped subsets. Each model contained is the same with $T_{\theta}$ and initialized randomly for its parameter. Then the networks will be trained separately using the corresponding subsets $D_{1\pi}^k(s), D_{2\pi}^k(s),...,D_{n\pi}^k(s)$:
\begin{equation}
T_{\theta_n} \leftarrow f(D_{n\pi}^k(s))
\end{equation}
where $f$ denotes the training function. Each subset contains the same amount of data units as the original $D_{\pi}^k(s)$, but each trajectory in these subsets is uniformly sampled from the original dataset. According to bootstrapped principle, the corresponding parameters $\theta_1, \theta_2, ... , \theta_n$ can be seen as $n$ independent and identically distributed (i.i.d.) samples from $P(\theta, D_{\pi}^k(s))$. The set  $\mathcal{T}$ (described in Eq. \eqref{transition_set}) is then approximated using the $n$ transition models:

\begin{equation}
\mathcal{T}=\{T_{\theta_1}, T_{\theta_2}, ... T_{\theta_n}\}
\end{equation}

The outputs of the environmental transition models indicate the potential future transitions of surrounding vehicles. According to the bootstrapped principle, the variance of $T_{\theta_1}, T_{\theta_2}, ... T_{\theta_n}$ outputs will be high when the data is insufficient to estimate the true transition confidently. In this case, the diverse outputs of transition models indicate the possible unexpected behavior of surrounding agents. Notably, each transition model $T_{\theta_n}$ considers the agent's stochastic in future behavior, commonly called aleatoric uncertainty. The constructed transition confidence set captures the epistemic uncertainty due to the inability to estimate the transition accurately. Intuitively, if the transition models in the set output significantly different results, one can consider that the surrounding agents may occupy a sizeable drivable space. 
Conversely, it should take up less space when the difference is small. In this study, the size of the occupied area can be regarded as a representation of the conservative level, but not explicitly estimated and used. Instead, the transition confidence will be passed directly to the planning stage, as described in the next section. In addition, the outputs of these models are constrained within the reachable set, which is determined by the agent’s dynamics \cite{althoff2021set}.

We use a Gaussian neural network as an example to build the transition model $T_{\theta}$ \cite{chua2018deep}:
\begin{equation}
T_{\theta}\left(s_{t+1} \mid s_{t}, {a}_{t}\right)=\mathcal{N}\left({\mu}_{\theta}\left(s_{t}, {a}_{t}\right), \boldsymbol{\sigma}_{\theta}\left({s}_{t}, {a}_{t}\right)\right)
\end{equation}
where $\mu$ and $\sigma$ denote the mean and variance of Gaussian distribution $\mathcal{N}$, respectively. 
This network can train supervised with a data unit $\{s_t, a_t, s_{t+1}\}$, which can be easily truncated from a trajectory $\omega_{\pi}$ in the dataset. 
Other probabilistic models, e.g., the generative model, can be used to build the transition model and easily constructed as an ensemble network. Moreover, the distribution indicating confidence comes from random initialization and the dataset resampling, which is independent of the network structure and easy to implement.

Notably, the number of transition models $n$ in the ensemble network structure will affect the performance of  $\mathcal{T}$ set estimation. Because the $n$ value determines the sample number from the distribution $ P(T_s(s), D_{\pi}^k(s))$,  $n$ should be sufficiently large to capture the  distribution character. Intuitively,  a larger $n$ value means the confidence set contains more samples from the distribution, and hence is more likely to cover the (approximate) true value. In other words, it means a stricter guarantee that the transition confidence set can cover true transition (Eq. \eqref{transition_set}).
In the experimental part, we will test the impact of different $n$ values on the set estimation and DCP performance. The overall process of ``long-tail'' rate estimation is shown in Algorithm 1.

\begin{algorithm}  
\caption{Ensemble Network Training and ``Long-Tail'' Rate Estimation}  
\LinesNumbered  
\KwIn{Policy $\pi$, Historical Driving Dataset $D_{\pi}^k(s)$,  Transition Model $T_{\theta}$}  
\KwOut{``Long-tail'' Rate $L=-Q_{\pi L}(s, \pi(s), D_{\pi}^k(s))$}  

Build ensemble network included  $T_{\theta_1}, T_{\theta_2}, ... T_{\theta_n}$  using transition model $T_{\theta}$

Sampled subdatasets $D_{1\pi}^k(s), D_{2\pi}^k(s), ... ,D_{n\pi}^k(s)$ from $D_{\pi}^k(s))$

Train each transition model in the ensemble network with sub-datasets independently

\For{ $T_{\theta_n}$ in ensemble network structure}
    {
    Generate several imaginary trajectories as Eq. \eqref{imaginary_trajectory}
    
    Estimate the driving policy performance using Eq. \eqref{mve}
            
    }

Calculate the ``long-tail'' rate using Eq. \eqref{long_tail_rate_set}

\end{algorithm}  

\subsection{Dynamically Conservative Planning Process}

The dynamically conservative planning process comprises two steps: generate candidate policies and choose the dynamically conservative action. 

\subsubsection{Candidate Policies Generation}

In each planning step, the DCP will generate a group of candidate policies $\Pi$ with different conservative levels:
\begin{equation}
\Pi = G_d(s)
\end{equation}
where  $s$ denotes the encountered driving case, and $G_d$ denotes the candidate policy generation function.

\begin{figure}[ht]
	\centering
	\includegraphics[width=0.95\linewidth]{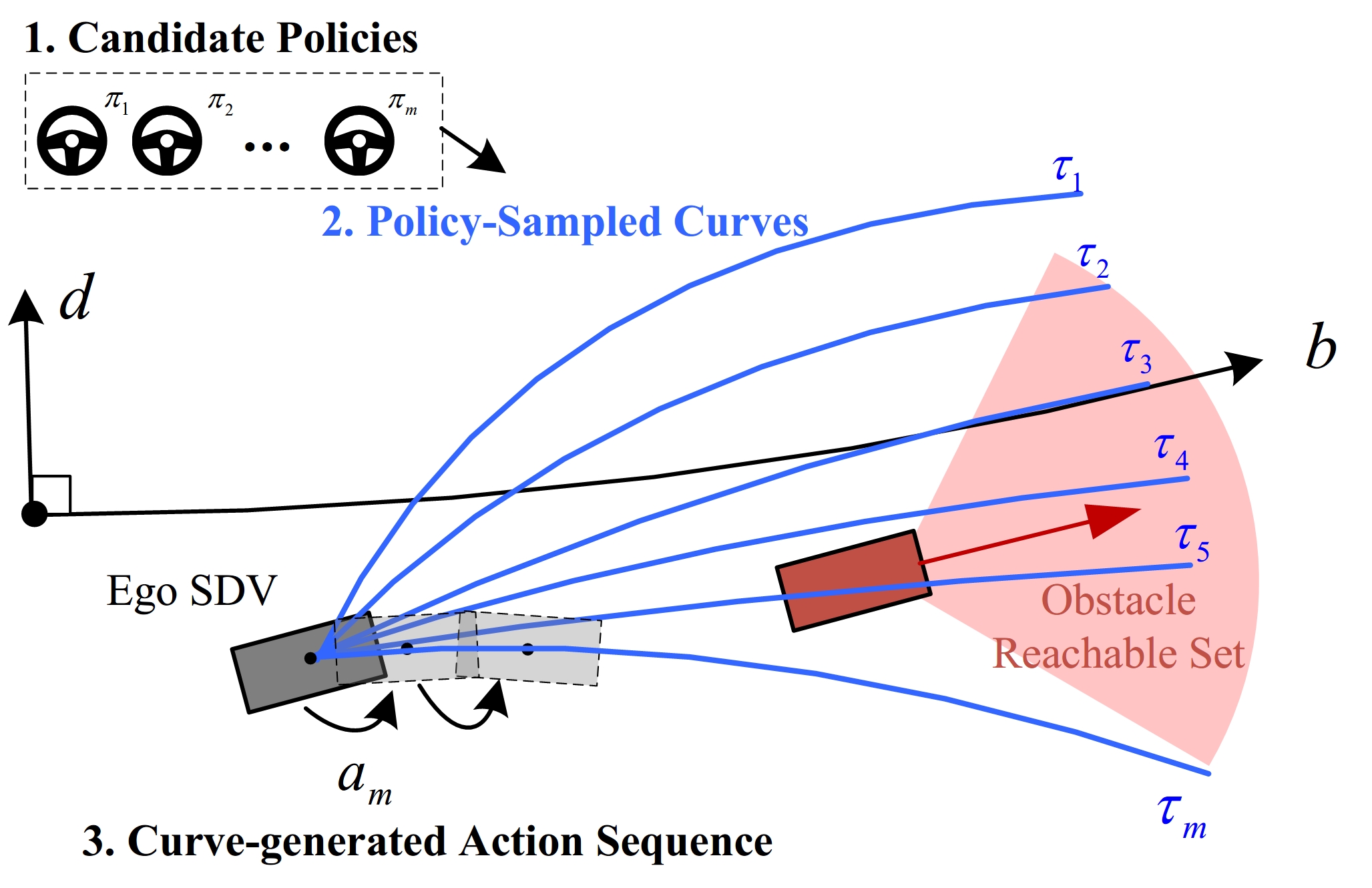}
	\caption{The candidate action generation process uses the reference path’s Frenét frame, i.e., $b$ denotes the driving distance along the reference path and $d$ denotes the deviation from the reference path. The DCP contains $m$ candidate policies. Each policy will generate a polynomial curve for the encountered driving state $s$. Each sampled curve is constrained by the ego SDV's target state after time horizon $H$, various in lateral movement and longitude velocities. Each curve can be transformed into an action sequence by making the ego SDV track the curve with a controller. The curves indicate different conservative levels, e.g., $\tau_1$ is more conservative than $\tau_3$ in the example case.}
	\label{fig:candidate_policies}
\end{figure}

We defined the action space as a discrete set containing $m$ candidate actions. 
\begin{equation}
a \in \mathcal{A}, A = \left(a_1, a_2, ..., a_m \right)
\end{equation}

The actions are generated using the candidate policies set in each planning step:
\begin{equation}
\Pi = \left(\pi_1, \pi_2, ..., \pi_m \right)
\end{equation}\begin{equation}
a_m = \pi_m\left(s\right)
\label{policy_to_action}
\end{equation}

The method to generate candidate actions through candidate policies is based on polynomial curve generation. Particularly, each policy $\pi \in \Pi$ will generate a fixed-length of curve $\tau$ for an encountered driving state $s$, which can be transferred to a fixed-horizon sequence of actions, beginning with $a_m$ (Fig. \ref{fig:candidate_policies}).  The curve is generated based on lattice planner  \cite{werling2010optimal} under the \frenet frame built on the reference path. The curve is a polynomial that starts from the ego SDV's current state $s^e_{0}=\left[p_{0}, \dot{p}_{0}, \ddot{p}_{0}\right]$ at $t=0$ and ends at a target state $s^e_{H}=\left[{p}_{H}, \dot{p}_{H}, \ddot{p}_{H}\right]$ at time $t=H$. $p_{0}$ and ${p}_{H}$ denote the ego SDV's positions under the \frenet frame at time $t=0$ and $t=H$, respectively. 

Such a quintic polynomial is proved optimal for cost function $C$ in the form of:

\begin{equation}
C=k_{j} J_{t}+k_{t} g(H)+k_{p} h\left(p_{H}\right)
\label{equ:cost1}
\end{equation}
where $J_t$ denotes the time integral of the square of jerk, which commonly indicates the comfort of ego SDV follows the curve:
\begin{equation}
J_t=\int_{t=0}^{H} \dddot{p}^2(\tau) d \tau
\label{jerk}
\end{equation}
where $g$ is a function of curve length $H$, fixed in this study. $h$ is the function of the target state . $k_{j}, k_{t}, k_{p}>0$. 
Each curve denotes a sequence of ego SDV's expected states in time horizon $H$ starting from the current encounter state. 

Each curve can be transformed into an action sequence of ego SDV's actions in time horizon $H$. It is achieved by making the ego SDV follow the curve with a controller. In this work, we use the preview controller that has an acceptable bounded tracking error \cite{xu2019design}. As such, a policy can be used to generate a trajectory with the action sequence described in Eq. \eqref{trajectory} for policy evaluation (Eq. \eqref{mve}).

Different candidate policies vary in their target states to generate different conservative levels. The policies' target states are sampled with different lateral movements and longitude velocities under the \frenet frame. The conservative level is defined according to how much the surrounding obstacles are enlarged. The planner with different conservative levels will adopt different generated curves/policies. Considering Fig. \ref{fig:candidate_policies} as an example, a not conservative planner may choose $\tau_4$ or $\tau_5$ considering the obstacle's estimated transition (red arrow).  A more conservative planner will choose $\tau_2$ or $\tau_3$ to avoid potential unexpected behavior of the obstacle. The most conservative planner considering the obstacle's reachable set will choose $\tau_1$ in the case. Thus, the generated curves indicate different conservative levels.

\subsubsection{Dynamically Conservative Action Generation}

\begin{figure}[ht]
	\centering
	\includegraphics[width=\linewidth]{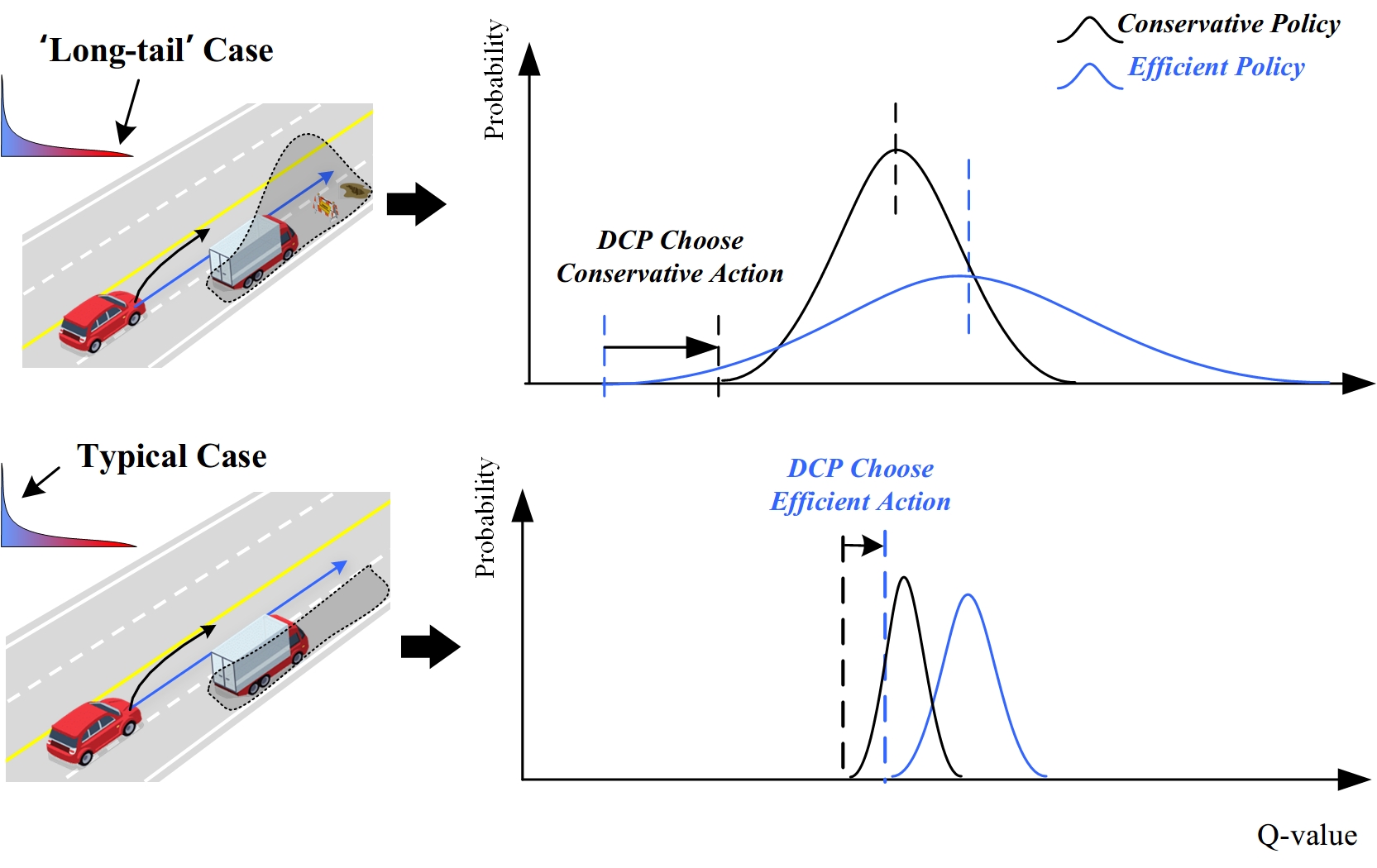}
	\caption{The principle of the dynamically conservative planning process. In low-confidence cases, an efficiency policy has higher performance expectations but a potentially higher probability of low-performance ``long-tail'' events. The conservative action has a higher performance lower bound when encountering low-confidence cases. The DCP method will choose the policy with a higher ``long-tail'' rate value $Q_{\pi L}(s, \pi(s), D)$, thereby dynamically adjusting the conservative level in different cases.}
	\label{fig:conservative_relationship}
\end{figure}

The previous section describes the design of policy sets with different conservative levels. The ``long-tail'' rate of each policy $L_{\pi} (s,D)$ can be estimated using the collected historical training data by the methods introduced in the previous section. Then, the dynamically conservative action is chosen by minimizing the ``long-tail'' rate $L_{\pi} (s,D)$ (maximizing the $Q_{\pi L}(s, \pi(s), D)$ value):

\begin{equation}
\begin{split}
a_{dc} =& \arg \min_{\pi \in \Pi} L_{\pi} (s,D) \\ 
=&\arg \max_{\pi \in \Pi} Q_{\pi L}(s, \pi(s), D) \\
=&\arg \max_{\pi \in \Pi} \min_{T_s  \in \mathcal{T}}\hat{Q}_{\pi}\left(s, a_{\pi}, T_s\right)
\end{split}
\label{dynamically_conservative_action}
\end{equation}

In this way, the DCP chooses the action that is least likely to cause potential safe-critical ``long-tail'' events, i.e., the policy with the highest performance lower bound considering the confidence of transition estimation. Eq. \eqref{dynamically_conservative_action} naturally yields dynamically conservative decisions in different cases. A typical example is shown in Fig.  \ref{fig:conservative_relationship}. In typical cases, the DCP can better estimate the environmental transition because of sufficient training data; thus, each candidate policy's ``long-tail'' rate is close to its true performance. Then, the DCP is more likely to choose the optimal policy that is more efficient. In ``long-tail'' cases, there is insufficient data to estimate the transition and guarantee the performance of the optimal policy. In addition, the conservative policy is more likely to have a higher performance lower bound. Another more straightforward explanation is that the transition confidence in the ``long-tail'' case is lower, making the surrounding agent's future likely to occupy more driving space. Thus, the ego SDV needs to make more conservative decisions to avoid possible collisions. The overall process of dynamically conservative planning is shown in Algorithm 2.

Considering the ``worst-case” may lead to very conservative behaviors (e.g., deadlocks that the ego SDV cannot move), but is necessary in ``long-tail” cases, as other more efficient policies cannot guarantee performance. In addition, the ego SDV keeps updating the policy according to the estimated confidence (``long-tail” rate). Therefore, it is not always conservative but can move when the planner has enough confidence to perform an efficient action.

The reward function $R$ should include two parts: the driving efficiency reward $r_e$ and safety reward $r_s$:

\begin{equation}
r = r_e + r_s
\label{reward}
\end{equation}

Notably, the proposed method has no requirements for the specific form of the reward function. 
The conservative concept can be broader and refers to policies with lower performance expectations but higher lower bound guarantees. For example, when the reward function includes both driving efficiency and safety terms, conservative policies tend to be inefficient but naturally have a higher safety lower bound. 
The reward function can also consider other aspects, such as driving comfort, local traffic rules, and passenger preference.

The DCP theoretically has $n$ times of computational cost as $n$ transition models are used.  Though not explored in this study, the $n$ models can be designed to work in parallel for acceleration in the future. More specifically, each model $T_s$ is used to calculate the value $\hat{Q}_{\pi}\left(s, a_{\pi}, T_s\right)$ in parallel. The parallel design will not affect the planner's performance. In this way, the proposed method can work with potentially larger $n$ values and not reduce calculation efficiency significantly (compared with the baseline). The real-time performance of the DCP will be studied in a case study.

\begin{algorithm}  
\caption{Dynamically Conservative Planning Process}  
\LinesNumbered  
\KwIn{Encountered Driving State $s$}  
\KwOut{Dynamically Conservative Action $a_{dc}$}  

Generate candidate policies with different conservative levels $\Pi = \left(\pi_1, \pi_2, ..., \pi_m \right)$

\For{ $\pi_m$ in $\Pi $}
    {
    Estimate the policy's ``long-tail'' rate $Q_{\pi L}(s, \pi_m(s), D)$ using Algorithm 1
    }

Generate dynamically conservative action $a_{dc}$ using Eq. \eqref{dynamically_conservative_action}

\end{algorithm}  

\section{Case Study}

\subsection{Experiment Setting}

\subsubsection{Simulation Scenario Design}

We design a typical road scenario in the CARLA simulator for the experiment \cite{dosovitskiy2017carla}, as shown in Fig. \ref{carla}. In this scenario, the ego SDV needs to complete an unprotected left-turn driving task starting from the top right corner of the intersection. We generate a group of driving cases in the scenario, which vary in initial physical status and driving intention of surrounding agents. The driving intention is designed as an agent's goal of driving, i.e., turning left or right.
An episode for the driving case starts by initialing all agents and will end when the ego SDV finishes the task, collides, or stays still for a long time. 

During the test, the ego SDV should plan the trajectory for each episode, and a preview controller will track the planned trajectory \cite{xu2019design}. The generated surrounding agents will react to the ego SDV using the driving model built in the simulator. The simulation runs at 10 Hz, the same as the ego SDV's decision frequency.

\subsubsection{Training and Testing Cases Generation}

We design training and test datasets based on the scenario to simulate ``long-tail'' problems in reality. More specifically, we generate 300 driving cases based on the designed scenario by randomly initializing the status and driving intention of surrounding agents. The randomized parameters include the surrounding vehicles’ initial position, initial velocity, and driving intention. The candidate initial positions are limited to within the drivable area of the road. The initial distance between two vehicles should be larger than five meters. The initial velocities range from $[0, 20$ km/h$]$. 

\begin{figure}[ht]
    \centering
    \includegraphics[width=0.95\linewidth]{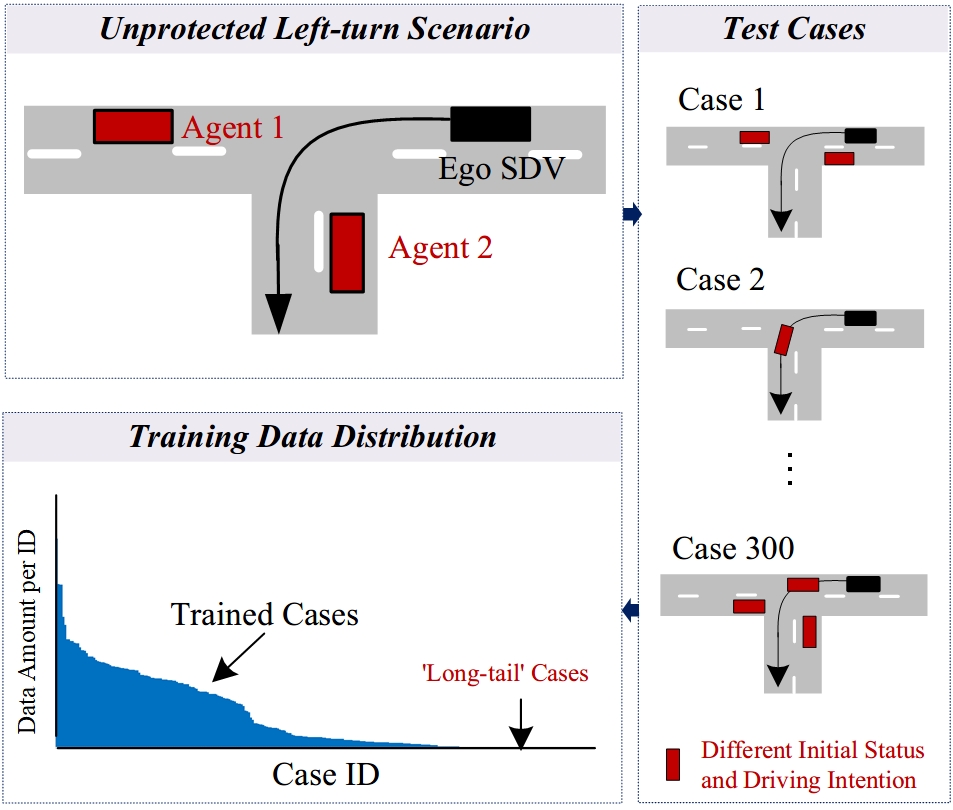}
    \caption{The test scenario built in CARLA. The ego SDV (blue) takes the black line as the reference path and aims to complete an unprotected left turn task. The surrounding agents (red) are set with random initial positions/velocities and driving intentions to generate 300 test cases. The training data amount of all test cases is distributed in the form of ``long-tail''. There include ``long-tail'' cases with zero training data.}
    \label{carla}
\end{figure}

During training, the proposed planner can collect driving data in all the cases but with different accessible amounts, as shown in Fig. \ref{carla}. The accessible training data amount for each case is defined as the number of episodes for which the ego vehicles can collect data. Zero data volume means the ego vehicle's planner has not seen the case during training. Then the planner will finally be tested in all 300 cases. In this way, the ego vehicle will encounter test cases that: 1) have already collected sufficient training data and 2) are rarely seen or unseen during training.

\subsection{Performance Metrics}

The performance metrics in the experiment include the following:

\subsubsection{Accuracy of ``Long-Tail'' Rate Estimation}
The accuracy of ``long-tail'' rate estimation is indicated by whether the following inequality is satisfied:
\begin{equation}
\tilde{Q_{\pi}}\left(s, a_{\pi}\right) \geq Q_{\pi L}(s, \pi(s), D_{\pi}^k(s))
\label{long_tail_accuracy}
\end{equation}

The inequality describes the minus of the ``long-tail'' rate as the lower bound of true driving performance. The true driving performance $\tilde{Q_{\pi}}\left(s, a_{\pi}\right) $ will be tested alone and estimated using the Monte Carlo policy evaluation method.

\subsubsection{Safety of Planner}
The safety of a planner in a driving case $s$ is defined as its non-collision rate:

\begin{equation}
P_{s}(s) = \frac{Y_s(s)}{Y(s)}
\end{equation}
where $Y(s)$ denotes the total number of test episodes for a case $s$ and is set to be 50. The planner will be tested for $Y(s)$ episodes in each case to evaluate performance under the stochastic test environment. 
$Y_s(s)$ denotes the number of episodes that the ego SDV can safely finish the task for each case. Correspondingly, the overall safety metrics of the planner for all cases is the average of $P_{s}(s) $ of all test cases.

\subsubsection{Efficiency of Planner}
The driving efficiency of a planner in a driving case $s$ is quantified by the ego SDV's average velocity:

\begin{equation}
P_{e}(s) =  \frac{\sum_{t=0}^{H} v_e(s,t)}{H}
\end{equation}
where  $ v_e(s,t)$ denotes the velocity of ego SDV at time $t$ during a test episode of driving case $s$. $H$ denotes the considered decision horizon. Correspondingly, the overall efficiency of the planner for all cases is the average of $P_{e}(s) $ of all test cases.

\subsection{Fixed Conservative Level Baseline Setting}
The baseline is designed to have a fixed conservative level in all driving cases. The state space, action space, reward function, and other parameters and settings of this baseline are the same as the proposed DCP in the tests (described in the next section). Thus, the baseline's task is also to choose an action from the candidate action set, the same as the DCP method. We design baselines with different fixed conservative levels. The conservative level is indicated by how much the space occupied by obstacles is inflated with prediction time. The most conservative baseline will be the reachable-set-based planner \cite{althoff2021set}, where the surrounding obstacles' all possible future are considered and avoided. The most nonconservative baseline will not enlarge surrounding obstacles. We designed a typical RL-based planner as this efficient baseline, which picks the action with the highest expected driving performance (described in Eq. \eqref{equ:Qdefine}) in all driving cases. The DCP method equals the efficient baseline when the number of transition models $n=1$.

\subsection{DCP Training Setting}

The proposed DCP is trained using the ``long-tail'' distributed data shown in Fig. \ref{carla}. The collected data will train the ensemble of transition models in the DCP structure. The state space, action space, and reward function design are described as follows.

\subsubsection{State Space}
The state-space used in our experiments comprises the states of the ego SDV and surrounding road users' information:
\begin{equation}
s \in \mathcal{S}, s=\left\{s_{e}, s_{1}, s_{2} \ldots s_{i}\right\}
\end{equation}
where $s_e$ denotes the ego SDV state, including position, pose, and velocity. Similarly, $s_i$ refers to the surrounding road users' states with the same information. The quantity of surrounding agents varies in real driving. Because the input dimension needs to be fixed for the used neural network that builds the transition model, we manually set the maximum number of obstacles as the network input size and create blank obstacles if the actual agent number is less. The blank obstacles will be created at a far distance away from the ego SDV with static status. We take the nearest four objects for simplification in the experiment. Though not explored in this study, graph neural networks can be used to build transition models with varying input dimensions for the DCP \cite{salzmann2020trajectron++}.

\subsubsection{Action Space}

The action space is generated using the candidate policies with different conservative levels in each planning step, as described in Eq. \eqref{policy_to_action}. We design 10 candidate policies in the experiment. According to Section III-B-1, each policy is determined by a target state of the ego SDV after horizon time $H$. We sample 9 target states with different lateral offsets and the longitudinal target velocities under the \frenet frame. The parameters for offsets and target velocities are the same in different cases (e.g., 15 km/h). In addition, the action space contains a brake action that makes the ego SDV stop as soon as possible. The candidate policies can generate curves with different expected velocities and offsets, yielding different conservative level driving.

\subsubsection{Reward}
The reward function described in Eq. \eqref{reward} is designed in detail. The driving efficiency reward $r_e$ encourages efficient and comfortable driving, which is designed similarly to Eq. \eqref{equ:cost1}:
\begin{equation}
r_e(s_t,a) = -k_{j} J(s_t,a)-k_{p} h\left(s_t\right) - k_v g(s_t)
\end{equation}
where $k_{j}, k_{p}, k_v>0$, and $J(s_t, a)$ is the function to calculate the ego SDV's jerk from the state $s_t$ and action $a$, similar to Eq. \eqref{jerk}:
\begin{equation}
J(s_t,a)=\int_{t}^{t+\Delta t} \dddot{p}^2(\tau) d \tau
\end{equation}
where $h(s_t)$ denotes the ego SDV's lateral distance $d_e(s_t)$ at state $s_t$ under the \frenet frame:
\begin{equation}
h(s_t)=|d_e(s_t)|
\end{equation}
where $g(s_t)$ denotes the difference between the ego SDV's speed $v_e(s_t)$ at state $s_t$ and the pre-set target speed $v_{\text {target }}$:
\begin{equation}
g(s_t)=\left|v_e(s_t)-v_{\text {target }}\right|
\end{equation}

The safety reward $r_s$ is defined as follows:
\begin{equation}
r_{s}(s_t)=\left\{
    \begin{aligned}
   r_{c} & , & collision, \\
    0 & , & else.
    \end{aligned}
\right.
\end{equation}
where $r_c$ is the collision penalty value.

\subsubsection{Network and Training Parameters}

The ensemble of transition models contained in DCP is trained with the data described in Fig.  \ref{carla}. Each transition model $T_{\theta_n}$ in the ensemble structure is a Gaussian neural network. It takes environment state and candidate action as inputs and contains two linear hidden layers with 128 neurons.  Then two linear layers decode the mean and various of future obstacle statuses, respectively. Different numbers of transition models in the ensemble structure will be tested. The parameters of the DCP in the evaluation are shown in Table \ref{parameter}.

\begin{table}
\caption{Parameters of Proposed DCP}
\begin{center}
\begin{tabular}{ccc}
\toprule  %
Parameters& Symbol& Value\\
\midrule  %
Collision penalty & $r_c$ & -500 \\
Ego vehicle's target velocity & $v_{\text {target }}$ & 30km/h \\
Cost weights of comfort & $k_j$ &0.1 \\
Cost weights of offset & $k_p$ & 1.0 \\
Learning rate & $l_r$ & 5e-4 \\
Action space size & - &  10 \\ 
Planning frequency & -- & 10 Hz\\
Time step & $\Delta t$ & 0.1s \\

\bottomrule %
\end{tabular}
\end{center}
\label{parameter}       

\end{table}

\subsection{Results}

\subsubsection{``Long-Tail'' Rate Estimation Accuracy}

\begin{figure}[ht]
	\centering
	\includegraphics[width=\linewidth]{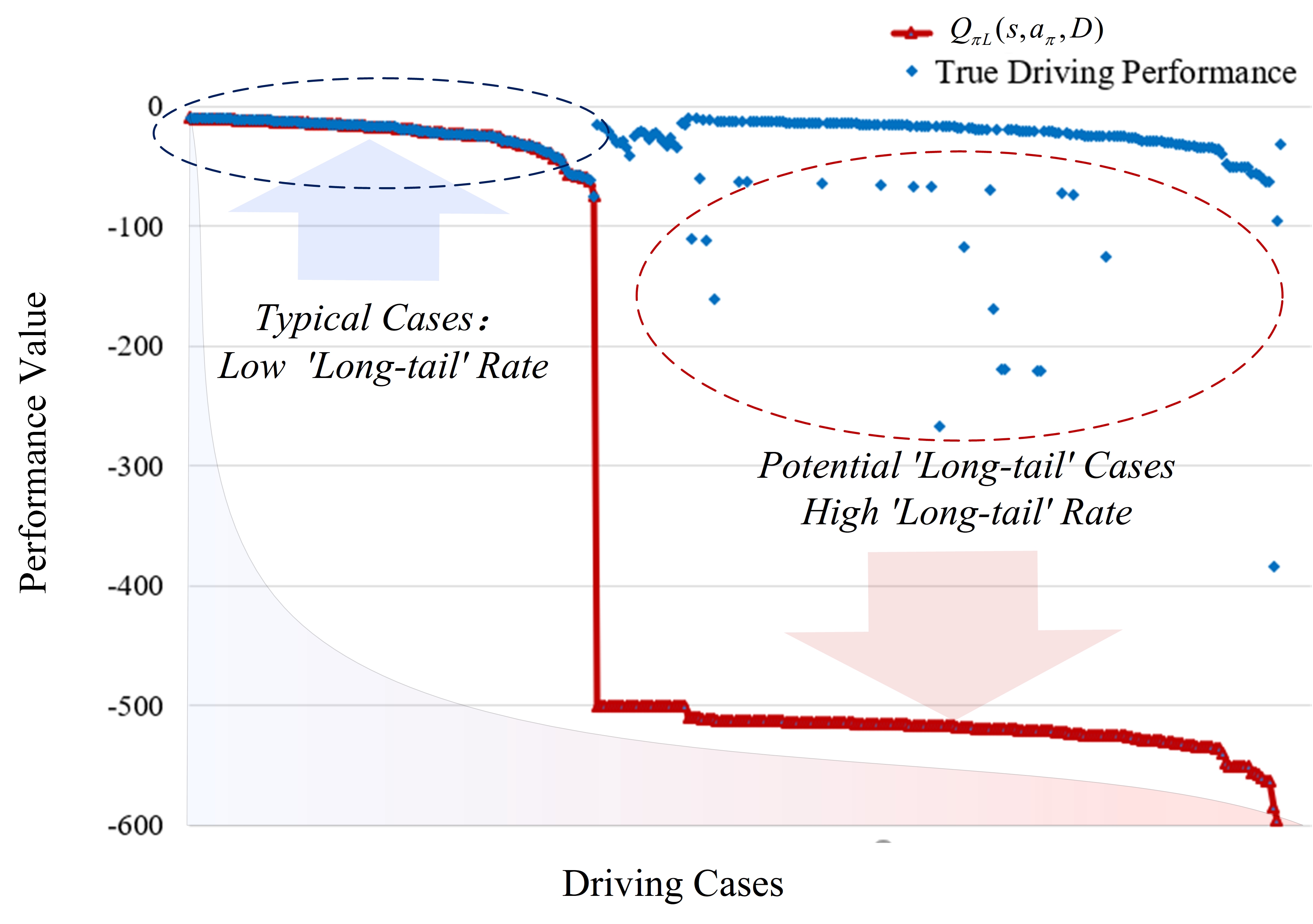}
	\caption{The accuracy of ``long-tail'' rate estimation in all driving cases. The driving cases are sorted by their ``long-tail'' rates from high to low. The results show that the ``long-tail'' rate is low in the potential low-performance cases (red circle) and high  in the typical cases (blue circle).  $Q_{\pi L}(s, a_{\pi}, D)$ is the lower bound of driving performance in all cases.}
	\label{fig:q_lower_bound}
\end{figure}

Fig.  \ref{fig:q_lower_bound} shows the estimated ``long-tail'' rate and the true driving performance in different driving cases. The true Q-value is estimated using the Monte Carlo policy evaluation method. The ego SDV is driving using the efficient baseline. 

The results show that the estimated ``long-tail'' rate is a lower bound of true driving performance in different driving cases. The rate is low in the potentially risky driving cases with low true driving performance (in red circle). The rate is high in the typical cases that the ego SDV can safely pass (in the blue circle). Notably, the proposed method does not achieve that by over-conservatively estimating the driving performance (i.e., designing a low ``long-tail'' rate in any case). In typical cases, the proposed  ``long-tail''  rate is a tight lower bound of true driving performance (blue circle).

\subsubsection{DCP Performance}

\begin{figure}[ht]
	\centering
	\includegraphics[width=\linewidth]{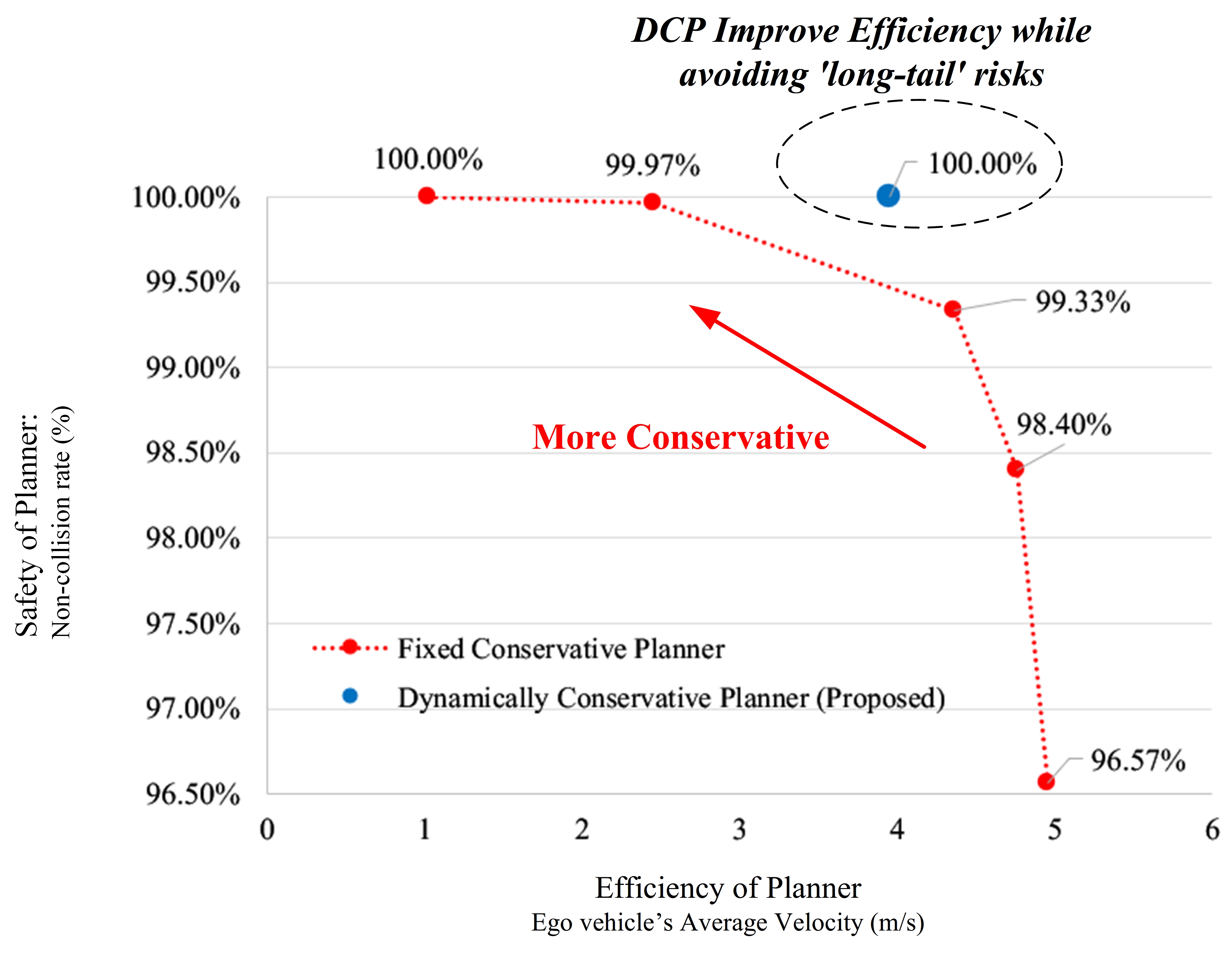}
	\caption{The overall performance of the DCP and baselines. The fixed conservative baselines tradeoff in driving safety and efficiency. A more conservative baseline is safer but has a lower average velocity. The proposed DCP has higher efficiency than the conservative baseline but simultaneously maintains the same safety level.}
	\label{fig:performance}
\end{figure}

\begin{table*}[!ht]
   \caption{Performance of DCP and baselines in different cases}
   \centering
    \begin{tabular}{ccccccc}
    \toprule
        \multirow{2}{*}{Planner} & \multicolumn{3}{c}{Safety of Planner (\%)} & \multicolumn{3}{c}{Efficiency of Planner (m/s)} \\
        ~ & Overall & ``Long-tail'' Cases & Typical Cases & Overall & ``Long-tail'' Cases & Typical Cases  \\ 
       \midrule  %
       Baseline (Efficient) & 96.57 & 93.97 & 100 & \textbf{4.953}   & \textbf{4.562} & \textbf{5.247}   \\ 
        Baseline (Conservative) & \textbf{100} & \textbf{100}  & \textbf{100}  & 1.019 & 0.798& 2.849  \\ 
        DCP &  \textbf{100} &  \textbf{100} & \textbf{100} & 3.951 & 1.571 & \textbf{4.947}   \\ 
    \bottomrule %

    \end{tabular}
    \label{table:performance}
\end{table*}

The performance of the DCP and fixed conservative baselines are shown in Table \ref{table:performance} and Fig. \ref{fig:performance}. 

Fig. \ref{fig:performance} shows the overall performance in all 300 test cases. The baselines with different fixed conservative levels need to balance driving safety and efficiency. The more conservative a baseline is, the lower the average velocity and the higher the safety rate it has. For example, the most conservative baseline makes safe decisions in all cases (100\% safety of planner) but has the lowest overall driving efficiency (1.019 m/s). Meanwhile, the most efficient baseline has a high driving efficiency (4.953 m/s) but makes some risky decisions (96.57\% safety). The proposed DCP method (with different $n$ values) has much higher efficiency than the conservative baseline while maintaining safety and guaranteeing performance in all cases. 

Table \ref{table:performance} shows the performance in different cases. We divided the test cases into ``long-tail'' and typical cases by setting a threshold of estimated ``long-tail'' rates. The results show that the conservative baseline and DCP can maintain safety in all cases, whereas the efficient baseline may make risky decisions in the ``long-tail'' cases. The driving efficiency of the conservative baseline is lowest in all driving cases (0.798 and 2.849 m/s for ``long-tail'' and typical cases respectively). In addition, the efficient baseline has a high average velocity in all cases. The proposed DCP method is conservative in ``long-tail'' cases (1.571 m/s) with high efficiency (4.947 m/s, close to that of the efficient baseline 5.247 m/s) in typical cases. The proposed DCP can adjust the conservative level according to the ``long-tail'' rate compared with the fixed conservative level baselines. As a result, it can significantly improve driving efficiency while ensuring safety.

\begin{figure}[ht]
	\centering
	\includegraphics[width=\linewidth]{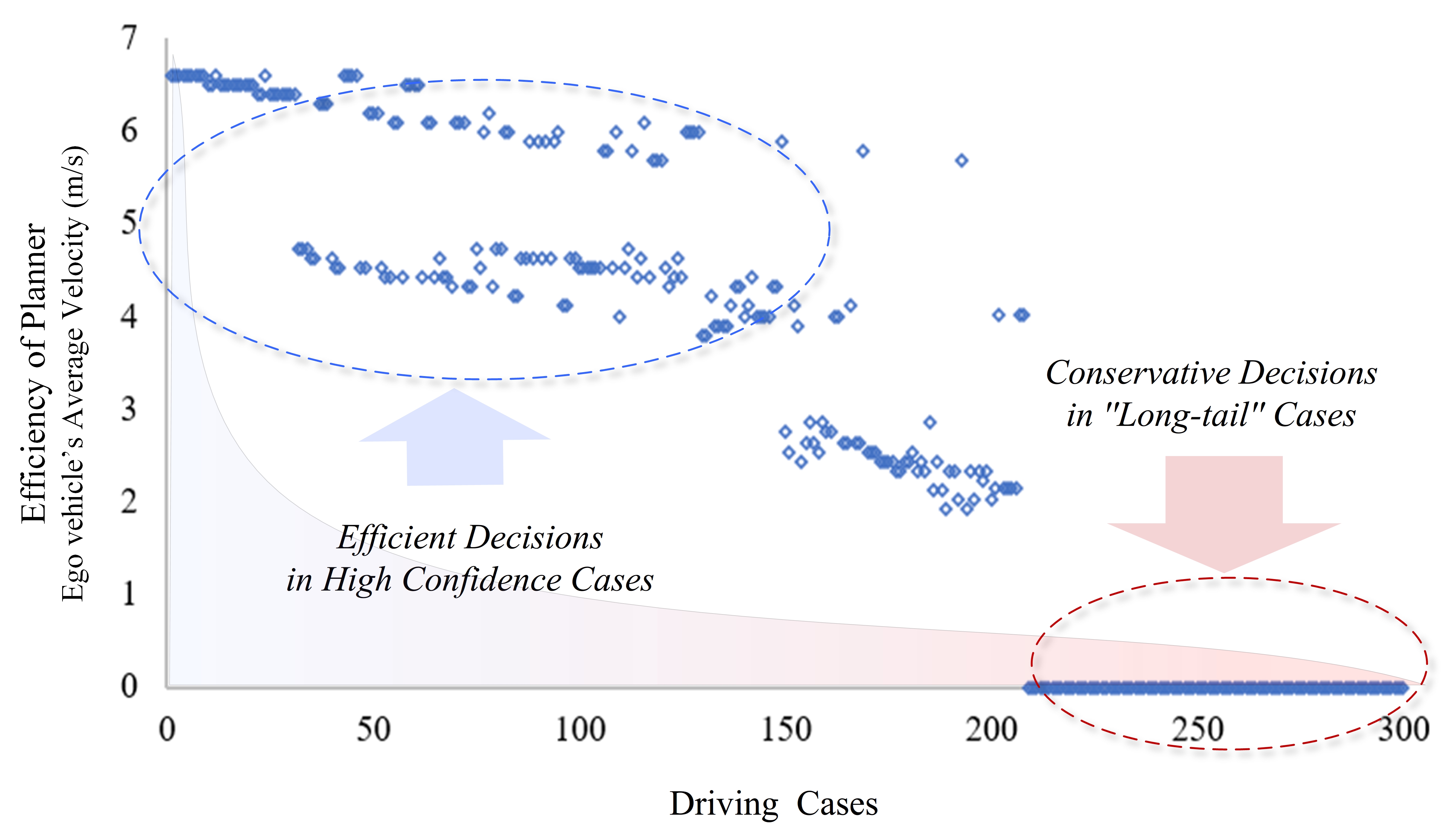}
	\caption{The driving efficiency of  the proposed DCP. The cases are sorted with the estimated ``long-tail'' rates from low to high. The result shows that the DCP can adjust the conservative level according to the estimated ``long-tail'' rate. It makes efficient decisions in typical cases while staying conservative in ``long-tail'' cases for safety.}
	\label{fig:velocity}
\end{figure}

The detailed analyses of safety and efficiency performance in each case are described as follows.
The planned velocity of the proposed DCP in all driving cases is shown in Fig. \ref{fig:velocity}. The driving cases are arranged in descending order of the ``long-tail'' rate. 
The results show that the proposed DCP can dynamically adjust the conservative level according to the ``long-tail'' rate. The ego SDV then has higher velocity in the high confidence cases, where the ``long-tail'' rate is high. Moreover, it is more conservative and plans lower speed when the ``long-tail'' rate decreases. The DCP outputs zero velocity (brake) in the potentially risky cases where the ``long-tail'' rate is low.

\begin{figure}[ht]
	\centering
	\includegraphics[width=\linewidth]{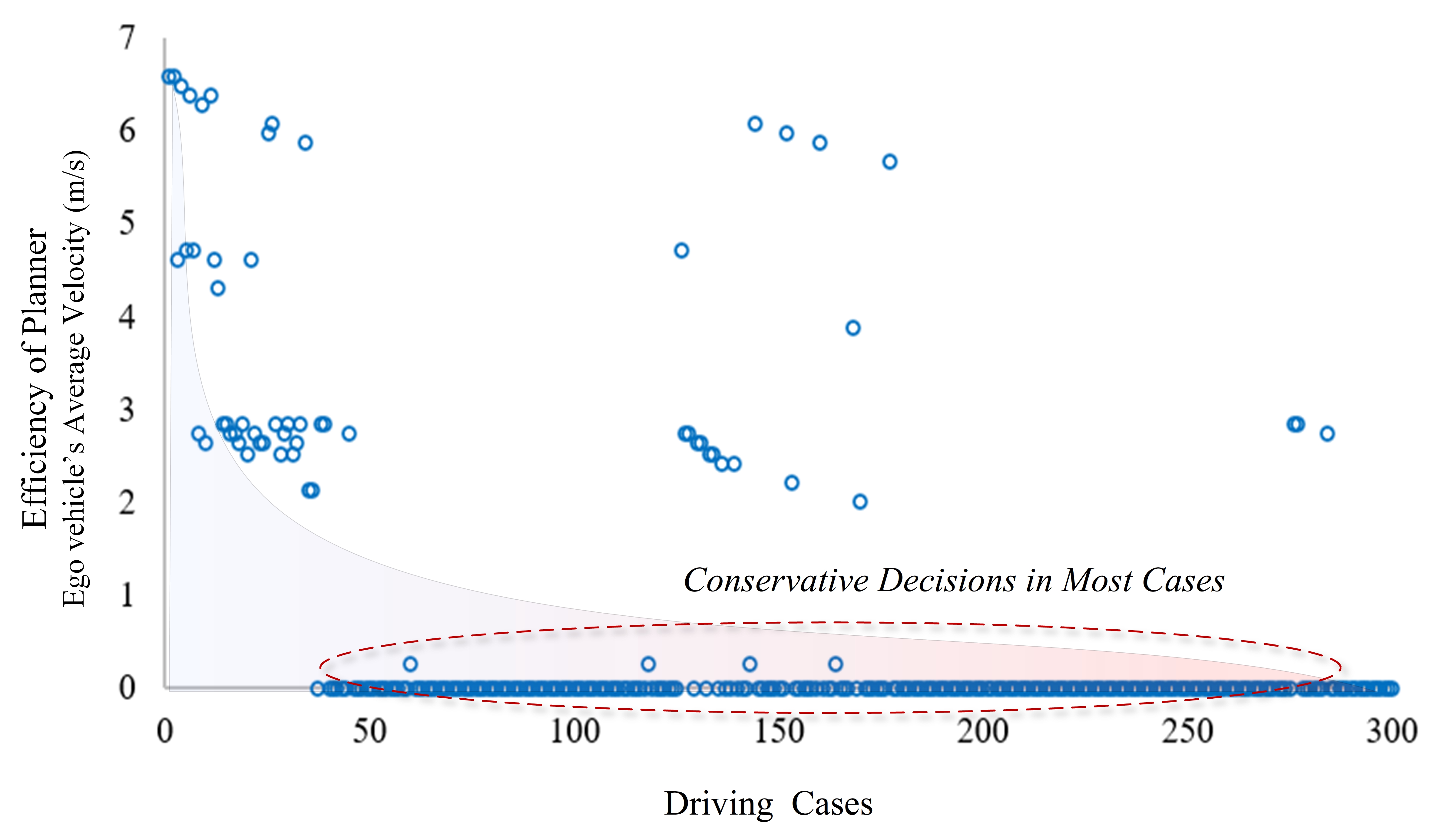}
	\caption{The driving efficiency of the most conservative baseline. It outputs low velocity (including braking) in most cases, making the ego SDV inefficient.}
	\label{fig:velocity2}
\end{figure}

\begin{figure}[ht]
	\centering
	\includegraphics[width=\linewidth]{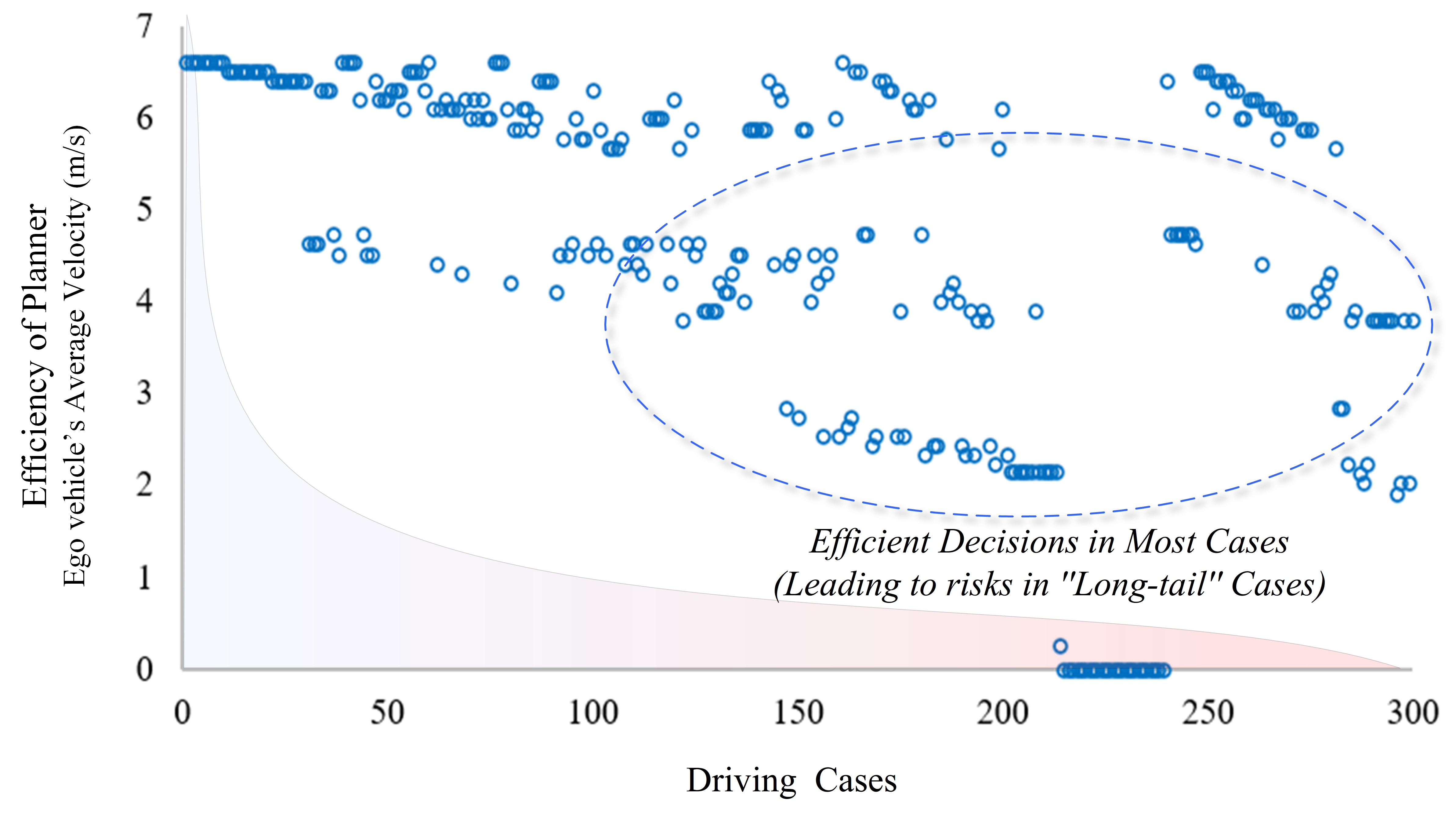}
	\caption{The driving efficiency of the least conservative baseline. It outputs high velocity in most cases and has a higher average velocity (than the conservative baseline). However, it may make risky decisions in ``long-tail'' cases, shown in Fig. \ref{fig:safety}.}
	\label{fig:velocity3}
\end{figure}

Notably, the proposed DCP dynamically adjusts the conservative level in different cases. Thus, its planned velocity is similar to that of the most efficient baseline in the typical cases where the ``long-tail'' rate and driving confidence are high, as shown by comparing the up-left side of Fig. \ref{fig:velocity} and \ref{fig:velocity3}. In addition, the DCP method acts like the conservative baseline in  ``long-tail'' cases and outputs brake actions. In conclusion, the proposed DCP can dynamically adjust the conservative level according to the estimated ``long-tail'' rate.

\begin{figure}[ht]
	\centering
	\includegraphics[width=\linewidth]{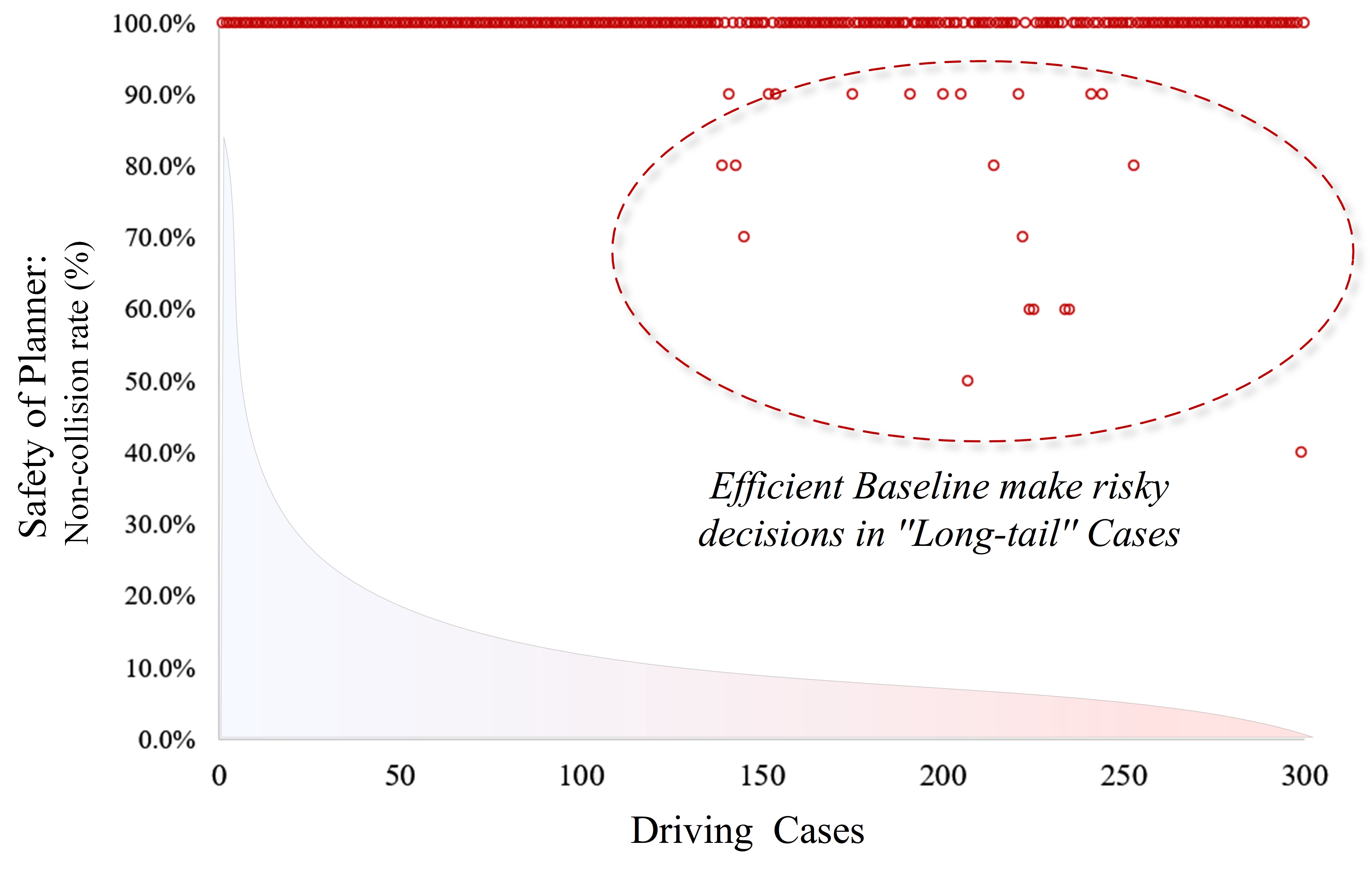}
	\caption{The safety performance of the least conservative baseline. It may make risky decisions, which lead to unsafe outcomes (in red circle). }
	\label{fig:safety}
\end{figure}

The safety performance of the efficient baseline is shown in Fig. \ref{fig:safety}. The driving cases are also arranged in descending order of the ``long-tail'' rates. The efficient baseline may fail in unexpected cases (red circle). It fails mainly in cases where the ``long-tail'' rate is low, and the ego SDV should have low confidence to pass, verifying the accuracy of the estimated ``long-tail'' rate. The conservative baseline is safe in all driving cases.

\subsubsection{The $n$ value's effect on DCP performance}

The number of ensemble models $n$ affects the DCP performance, as shown in Table \ref{Quantitative}. A larger $n$ means that the transition sets are more likely to cover the true transition. The results show that even two transition models have better performance (high safety and efficiency of planner) than the efficient baseline. $n=5$ is sufficient for the DCP to be safe in the experiment. When $n$ is larger than 10, the DCP performance will not be affected significantly in the test cases.

\subsubsection{Real-Time Performance of DCP}

The DCP is tested on a computer with Intel Xeon(R) E5-2620 CPU, 128 Gb RAM, and GeForce GTX 1080 Ti. The DCP (implemented with Python) runs at 10 Hz when the number of the transition models is twenty ($n$ is 20). Notably, though the DCP calculates trajectory/policy at 10 Hz, the downstream controller that converts the planned trajectory into control signals/actions (steering angle and brake) can work at a higher frequency to achieve real-time action updates in practice, e.g., 50 Hz. 

In conclusion, the proposed DCP improves driving efficiency to the same level as an efficient baseline in typical driving cases. In addition, the DCP maintains safety in ``long-tail'' cases where potential failures may occur by adjusting to being conservative. Therefore, the DCP significantly improves the overall driving efficiency over the conservative baseline. Further, the DCP avoids risks in low-confidence cases by accurately estimating the ``long-tail'' rate and adjusting accordingly. The result shows that the DCP can well balance conservation and efficiency according to the ``long-tail'' rate, validating the method's effectiveness.

\begin{table}[!ht]
\caption{DCP performance with different numbers of transition models $n$}
    \centering
     \setlength{\tabcolsep}{1mm}{
    \begin{tabular}{ccc}
    \toprule  %
      $n$ values & \makecell[c]{Safety of Planner \\$P_{s}$  (\%)  }  & \makecell[c]{Efficiency of Planner\\ $P_{e}$  (m/s)   }\\ 
        \midrule  %
       2 & 99.2 & 4.569 \\
     5 & 100 & 3.951 \\
    10 & 100 & 3.398 \\ 
    15 & 100 & 3.205 \\ 
    20 & 100 & 3.186 \\ 
        \bottomrule %
    \end{tabular}
    }
    \label{Quantitative}       
\end{table}

\subsubsection{Example Cases}

\begin{figure}[ht]
	\centering
	\includegraphics[width=\linewidth]{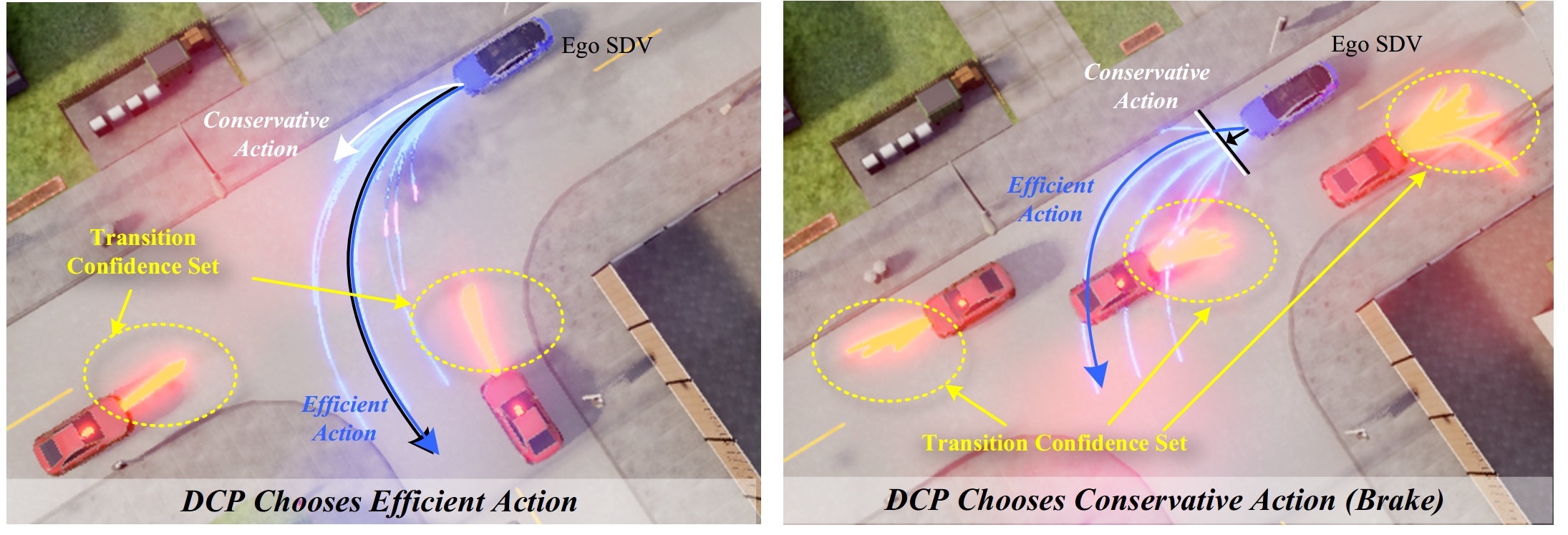}
	\caption{The two example cases recorded in the experiment. The blue vehicle is the ego SDV, and the red vehicles are surrounding agents. The candidate policies in the DCP generate the blue shine curves. The yellow lines are the surrounding agents' transition confidence set (only show the mean value of each transition). The black curve indicates the DCP action. The blue curve is the action of the efficient baseline, and the white curve is the conservative baseline's output action. The DCP chooses efficient policy in high-confidence cases (left) while being conservative to brake in potential ``long-tail'' cases (right).}
	\label{fig:cases}
\end{figure}

In this section, we describe two typical cases during the experiment, as shown in Fig.  \ref{fig:cases}. First, the left half shows the performance of the DCP in a high-confidence driving case, where the DCP is trained 218 times. 
In this case, the transition confidence set of surrounding agents is concentrated, i.e., the $n$ transition models have similar outputs. Thus, the transition confidence is high and leads to high-confidence in driving performance estimation (Eq. \eqref{q_t_relationship}). Therefore, in this case, the proposed DCP chooses an efficient action (as the efficient baseline). In the same case, the conservative baseline will select the action that considers the reachable set, i.e., decelerating and yielding to the road's right side.
The right half of Fig.  \ref{fig:cases} shows a low-confidence case, where 17 training episodes are accessible. The DCP has low confidence in the surrounding agents' transitions. The estimated transition confidence set is scattered. In this case, the DCP chooses a conservative action (brake), the same as the conservative baseline. The efficient baseline attempts to avoid the obstacle by keeping slightly to the right but fails and yields collisions.

\section{Conclusion}
In this study, we proposed a dynamically conservative planner (DCP) for ``long-tail'' cases. We formalized the ``long-tail'' problem as a driving confidence estimation problem and defined the ``long-tail'' rate for quantification. The DCP method automatically adjusts its conservative level according to the estimated ``long-tail'' rate. The DCP method is evaluated in the CARLA simulator. The results show that the DCP method can outperform baselines with a fixed conservative level in driving safety and efficiency. The DCP accurately estimates the ``long-tail'' rate of encountered driving cases. Thus, it adjusts to be conservative in potential ``long-tail'' cases and maintains efficiency in typical cases. Consequently, the DCP method has high efficiency while avoiding ``long-tail'' risks. 

This work contributes to solving the ``long-tail'' problem in safety-critical systems. Owing to the endless possibilities in the real world, all possible scenarios cannot be considered during the design process. This work defines ``long-tail'' as a confidence problem in statistics and refers to a conservative solution for possible risky cases. It provides a technique to guarantee a system's performance without resorting to a global conservative setting to avoid potential risks in all cases. Instead, the system will identify ``long-tail'' cases and adjust to being conservative. In the future, the ``long-tail'' problem in other SDV modules will be considered, e.g., perception, localization, and prediction. The DCP method will also be tested on real vehicles to demonstrate its performance.

\addtolength{\textheight}{1cm}   

\section*{ACKNOWLEDGMENT}

This work is supported by the National Natural Science Foundation of China (NSFC) (U1864203, 52102460) and China Postdoctoral Science Foundation (2021M701883)
It is also funded by the Tsinghua University-Toyota Joint Center.

\appendices

\ifCLASSOPTIONcaptionsoff
  \newpage
\fi

\bibliographystyle{IEEEtran}
\bibliography{ref.bib}

\begin{IEEEbiography}[{\includegraphics[width=1in,height=1.25in,clip,keepaspectratio]{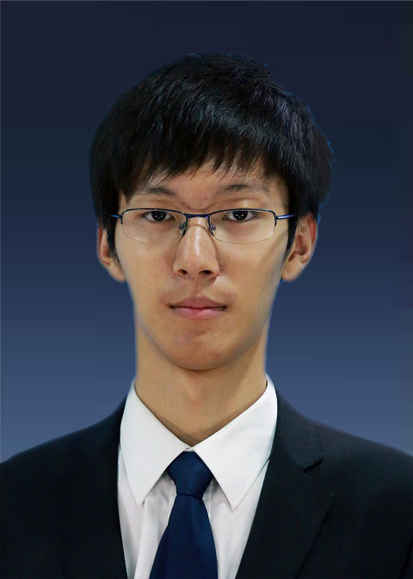}}]{Weitao Zhou} received the B.S. and M.S. in automotive engineering from Beihang University. 

He is currently a Ph.D. student at Tsinghua University. 
His research interests include autonomous driving, reinforcement learning and open-world learning.
\end{IEEEbiography}

\begin{IEEEbiography}[{\includegraphics[width=1.1in,height=1.25in,clip,keepaspectratio]{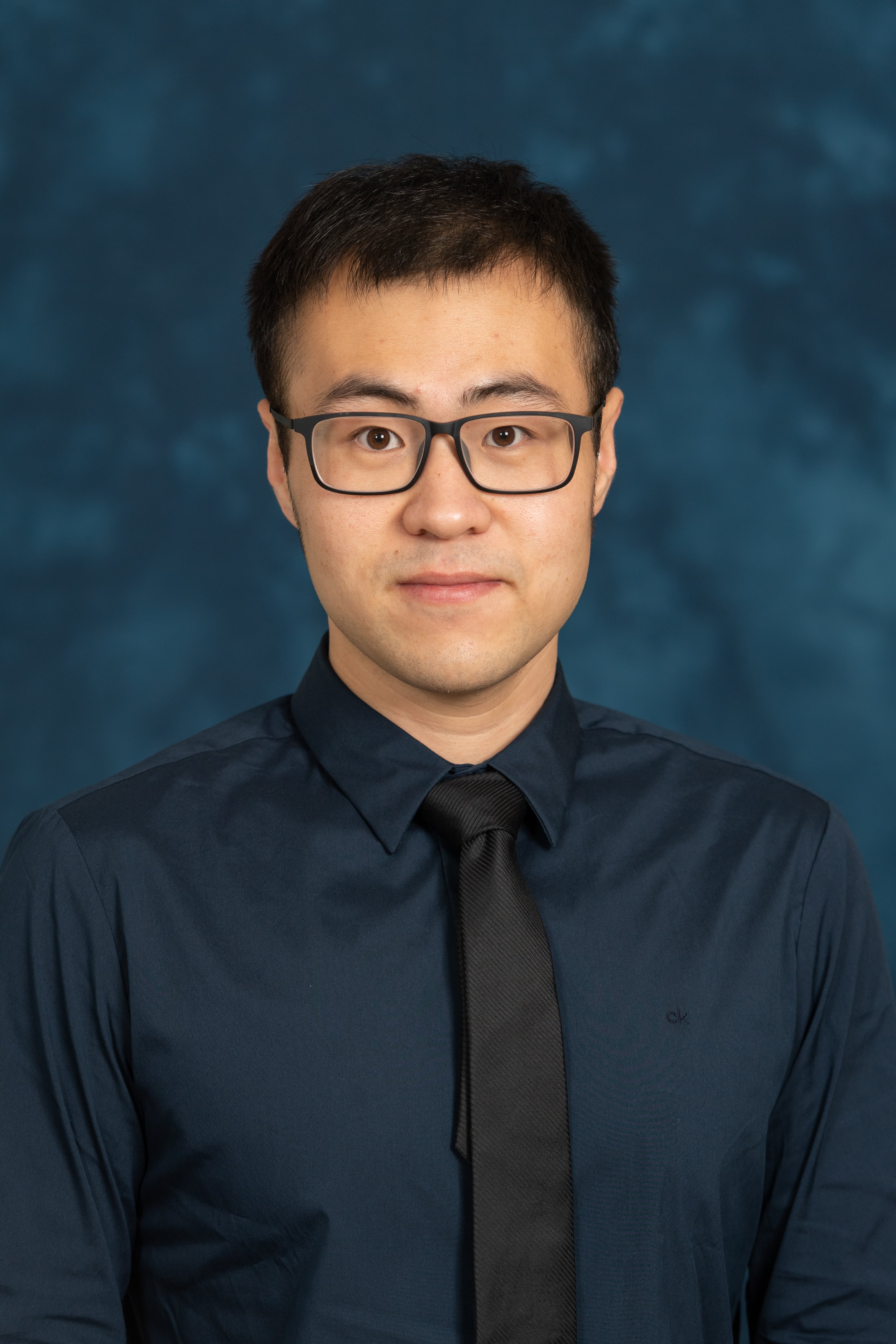}}]{Zhong Cao} received the B.S. and Ph.D. degree in automotive engineering from Tsinghua University in 2015 and 2020, respectively. 

He is currently a postdoc at Tsinghua University, with Shuimu Scholarship. 
His research interests include autonomous vehicle, trustworthy reinforcement learning, long life learning and HD map.
\end{IEEEbiography}

\begin{IEEEbiography}[{\includegraphics[width=0.9in,height=1.25in,clip,keepaspectratio]{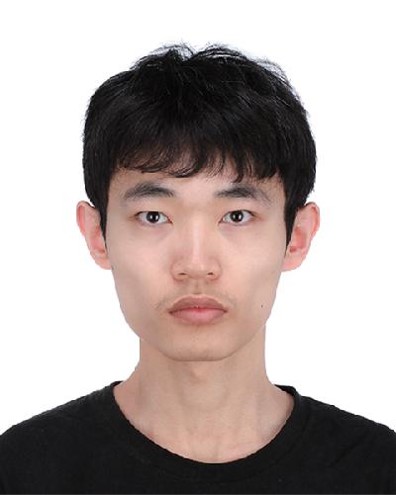}}]{Nanshan Deng} received the B.S. degree in automotive engineering from Tsinghua University in 2018. 

He is currently a Ph.D. student at Tsinghua University. His research interests include autonomous vehicle,  transfer reinforcement learning and meta learning.
\end{IEEEbiography}

\begin{IEEEbiography}[{\includegraphics[width=0.9in,height=1.25in,clip,keepaspectratio]{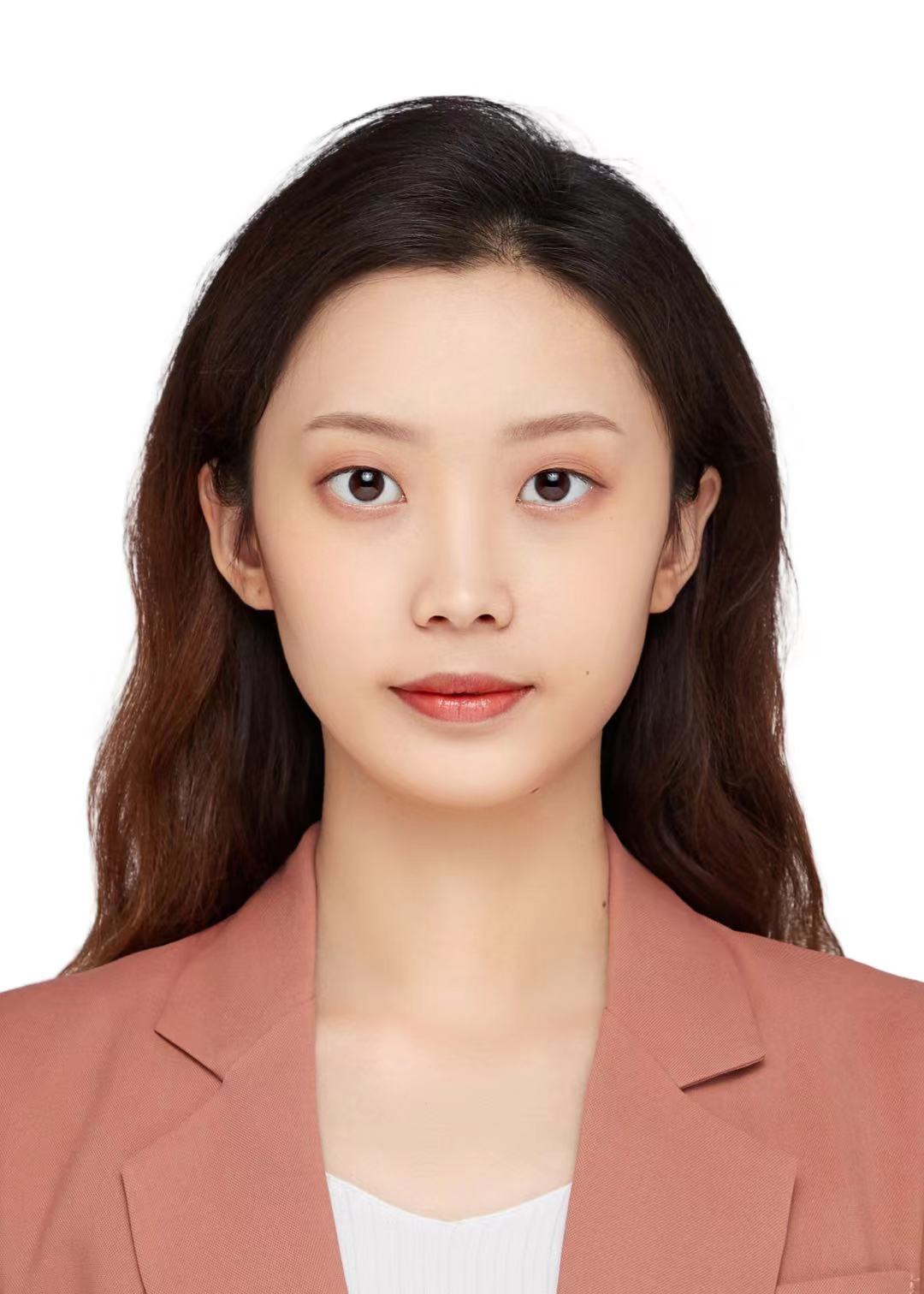}}]{Xiaoyu Liu} received the B.S. degree in automotive engineering from Tsinghua University in 2021. 

She is currently a master student at Tsinghua University. Her research interests include autonomous vehicle, simulation testing and corner case generation.
\end{IEEEbiography}
\begin{IEEEbiography}[{\includegraphics[width=0.9in,height=1.25in,clip,keepaspectratio]{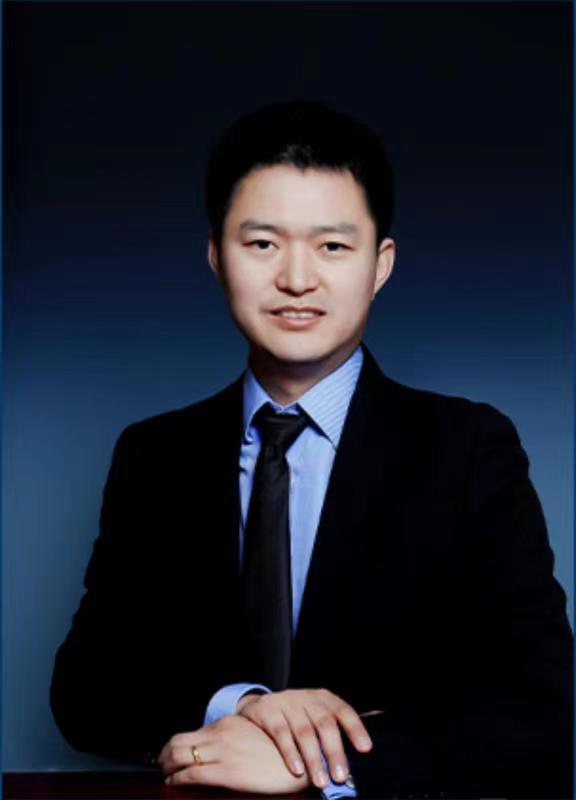}}]{Kun Jiang} received the B.S. degree in mechanical and automation engineering from Shanghai Jiao Tong University, Shanghai, China in 2011. Then he received the Master degree in mechatronics system and the Ph.D. degree in information and systems technologies from University of Technology of Compiègne (UTC), Compiègne, France, in 2013 and 2016, respectively. He is currently an assistant research professor at Tsinghua University, Beijing, China. His research interests include autonomous vehicles, high precision digital map, and sensor fusion.
\end{IEEEbiography}
\begin{IEEEbiography}[{\includegraphics[width=1.1in,height=1.25in,clip,keepaspectratio]{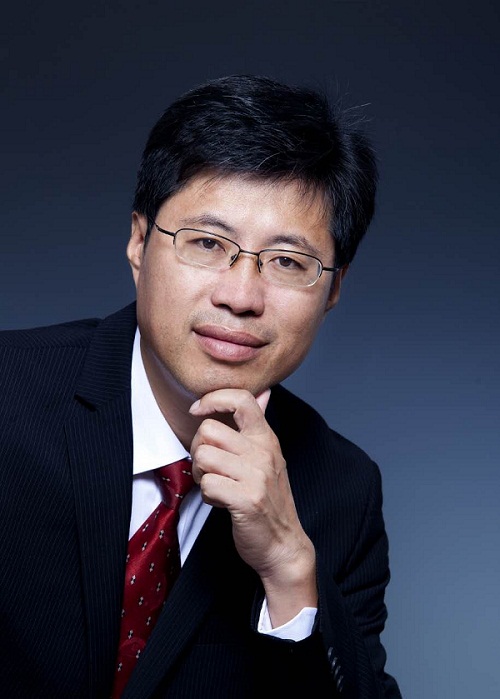}}]{Diange Yang} received the B.S. and Ph.D. degrees in automotive engineering from Tsinghua University, Beijing, China, in 1996 and 2001, respectively. He serves as the director of automotive engineering at Tsinghua university.

He is currently a Professor with the Department of Automotive Engineering, Tsinghua University. His research interests include intelligent transport systems, vehicle electronics, and vehicle noise measurement.

Dr. Yang attended in “Ten Thousand Talent Program” in 2016. He also received the Second Prize from the National Technology Invention Rewards of China in 2010 and the Award for Distinguished Young Science and Technology Talent of the China Automobile Industry in 2011.

\end{IEEEbiography}

\end{document}